\documentclass{article}

  \usepackage[main, final]{neurips_2025}

\usepackage[utf8]{inputenc} 
\usepackage[T1]{fontenc}    
\usepackage{hyperref}       
\usepackage{url}            
\usepackage{booktabs}       
\usepackage{amsfonts}       
\usepackage{nicefrac}       
\usepackage{microtype}      
\usepackage{xcolor}         

\usepackage{dsfont}     
\usepackage{amssymb}    
\usepackage{booktabs}
\usepackage[table]{xcolor}
\usepackage{amsmath}    
\usepackage{graphicx} 
\usepackage{algorithm, algorithmic}
\usepackage{wrapfig}

\title{MetaTPT: Meta Test-time Prompt Tuning for Vision-Language Models}

\author{%
  Yuqing Lei$^1$ \quad Yingjun Du$^2$ \quad Yawen Huang$^3$ \quad Xiantong Zhen$^4$ \quad Ling Shao$^1$ \\
  $^1$UCAS-Terminus AI Lab, University of Chinese Academic of Sciences\\
  $^2$AIM Lab, University of Amsterdam\\
  $^3$Jarvis Research Center, Tencent Youtu Lab\\
  $^4$Central Research Institue, United Imaging Healthcare Co., Ltd.
}

\begin{document}
\maketitle

\begin{abstract}

Vision-language models (VLMs) such as CLIP exhibit strong zero-shot generalization but remain sensitive to domain shifts at test time. Test-time prompt tuning (TPT) mitigates this issue by adapting prompts with fixed augmentations, which may falter in more challenging settings. In this work, we propose Meta Test-Time Prompt Tuning (MetaTPT), a meta-learning framework that learns a self-supervised auxiliary task to guide test-time prompt tuning. The auxiliary task dynamically learns parameterized augmentations for each sample, enabling more expressive transformations that capture essential features in target domains. MetaTPT adopts a dual-loop optimization paradigm: an inner loop learns a self-supervised task that generates informative views, while the outer loop performs prompt tuning by enforcing consistency across these views. By coupling augmentation learning with prompt tuning, MetaTPT improves test-time adaptation under domain shifts. Extensive experiments demonstrate that MetaTPT achieves state-of-the-art performance on domain generalization and cross-dataset benchmarks.

\end{abstract}
\section{Introduction}
\label{sec:intro}

Vision-language models (VLMs)~\cite{jia2021scaling, 
li2022blip, alayrac2022flamingo, yu2022coca} such as CLIP~\cite{radford2021learning} have exhibited strong zero-shot generalization. By pretraining on large-scale image-text corpora, CLIP learns a joint embedding space that aligns visual and textual representations. In zero-shot classification, an image is assigned to the class whose textual description—often instantiated using a template such as \textit{“a photo of a \{class\}”}—has the highest similarity with the image embedding. While this approach obviates the need for task-specific fine-tuning, zero-shot performance is contingent on the target domain following a similar distribution to its source domain. When there is a domain shift, its performance on the target domain will drop substantially.

To mitigate the impact of domain shifts, Test-Time Adaptation (TTA) have been proposed to enable models to self-adapt during inference. Instead of freezing the model after training, TTA updates certain model components by optimizing unsupervised objectives~\cite{wang2021tent,liang2020we,liang2021source} without access to labeled data from source domain. Among these, Test-Time Training (TTT)~\cite{sun2020test} introduces a self-supervised auxiliary task—such as image rotation prediction—optimized at test time, allowing the model to adjust its representations to new domains.

Building on this paradigm, recent work~\cite{feng2023diverse, abdul2023align,zhang2024historical,karmanov2024efficient,zhang2024boostadapter} such as Test-time Prompt Tuning (TPT)~\cite{shu2022test} extend TTA to VLMs, which adapts learnable prompts~\cite{zhou2022learning} on-the-fly while keeping the pretrained image and text encoders frozen. During inference, TPT employs fixed data augmentations~\cite{hendrycks2019augmix} to generate a batch of augmented views for each test sample, exposing the model to diverse visual variations during adaptation. The learnable prompts is then optimized through entropy minimization across these views, encouraging the model to produce confident predictions in the new domain.

\begin{figure}[t]
    \centering
    \includegraphics[width=\linewidth]{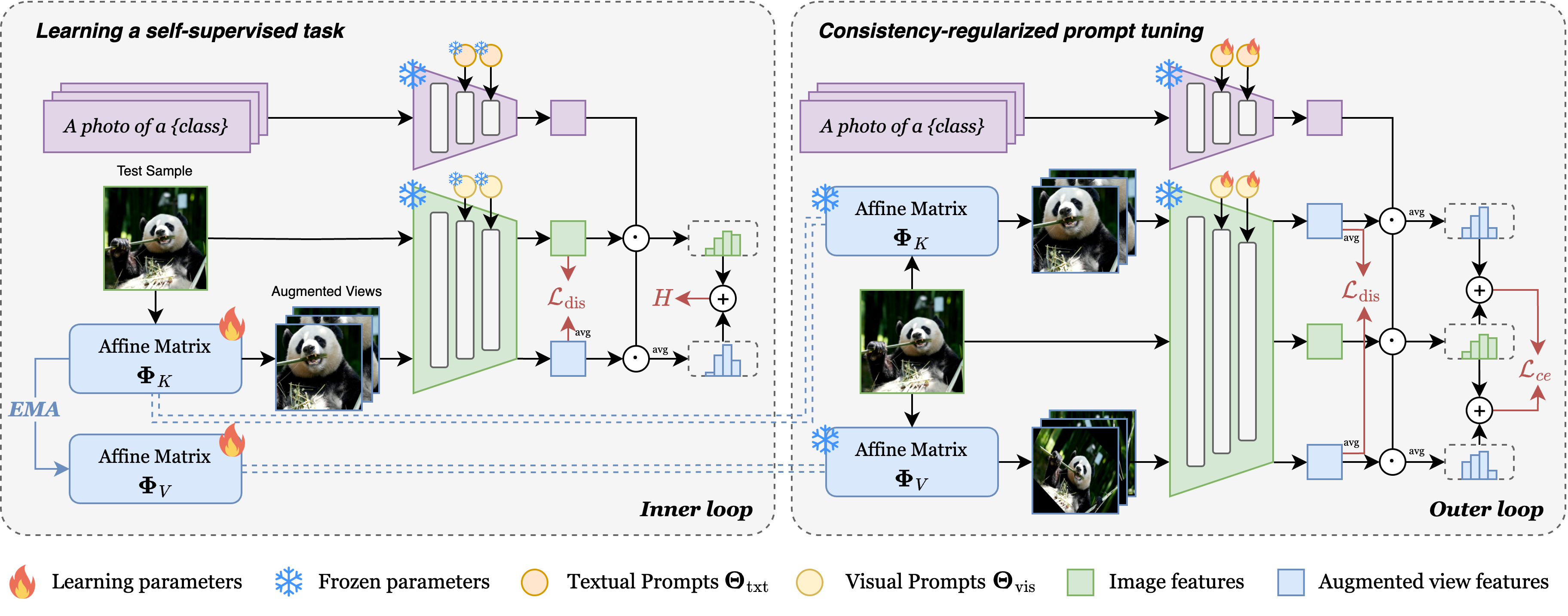}
    \caption{\textbf{Overview of MetaTPT}. This framework is formulated as a dual-loop meta-adaptation process. In the \textit{inner loop}, a self-supervised auxiliary task is learned to dynamically optimize parameterized augmentations: $\mathbf{\Phi}_K$ is updated through a joint objective combining $H$ and $\mathcal{L}_\textrm{dis}$ to produce informative views, while $\mathbf{\Phi}_V$ employs an EMA of $\mathbf{\Phi}_K$ to prevent collapse. In the \textit{outer loop}, the learnable prompts $\mathbf{\Theta}=\{\mathbf{\Theta}_\mathrm{txt},\mathbf{\Theta}_\mathrm{vis}\}$ by enforcing cross-view consistency, using a combined loss of $\mathcal{L}_{ce}$ and $\mathcal{L}_\textrm{dis}$ over the generated views.}
    \label{fig:pipeline}
    \vspace{-15pt}
\end{figure}

Inspired by TTT~\cite{sun2020test}, we propose an extension to TPT~\cite{shu2022test} that introduces a learnable self-supervised auxiliary task at test time. We observe the fix augmentations used in TPT may fail to capture the nuanced features necessary for discrimination, leading to unreliable adaptation especially in challenging target domains.
To overcome these limitations, the self-supervised auxiliary task is designed to dynamically learn augmentations during adaptation rather than manually defined. These augmentations are parameterized as differentiable affine transformations, enabling spatial operations—such as scaling, shearing, flipping, and translation.
This adaptability allows the model to capture complex, sample-specific variations such as subtle spatial distortions, shape deformations, or domain-specific artifacts that are common in challenging domains.

To effectively co-adapt parameterized augmentations with prompts, we formulate the adaptation as a bi-level optimization problem, naturally interpreted as a form of dual-loop meta-learning~\cite{finn2017model}. Building on this formulation, we introduce \textbf{Meta} \textbf{T}est-time \textbf{P}rompt \textbf{T}uning (\textbf{MetaTPT}) for vision-language models.
In this framework, the \textit{inner loop} optimizes a self-supervised task that dynamically generates adaptive augmentations, while the \textit{outer loop} updates the prompts by enforcing consistency across these views.
Following TPT~\cite{shu2022test}, our adaptation is performed online, with learnable parameters optimized for each test sample. Thus, MetaTPT serves as a meta-adaptation mechanism within a single test stream, where both loops operate per sample to improve generalization under domain shift.

Empirically, we evaluate MetaTPT on two zero-shot generalization benchmarks (domain generalization and cross-dataset evaluation) against four ImageNet-trained models (CLIP~\cite{radford2021learning}, CoOP~\cite{zhou2022learning}, MaPLe~\cite{khattak2023maple} and MMRL~\cite{guo2025mmrl}). Compared with TPT, MetaTPT demonstrates strong performance to domain shift and achieves more improvements on challenging datasets, highlighting the effectiveness of learnable augmentations in adapting to complex distribution changes.
Ablation studies further confirm the effectiveness of MetaTPT’s design.
First, compared to the fixed augmentations used in TPT, our learnable augmentations outperforms fixed augmentations all four base models, validating their generality across target domains.
Second, our dual-loop meta-learning framework (interleaved optimization of augmentations and prompts) outperforms one-stage joint optimization, emphasizing the benefits of meta-learning. 
Finally, our online adaptation strategy (meta-adapts augmentations and prompts per sample) surpasses an offline variant (meta-train augmentations across all samples before adapting prompts individually), underscoring the importance of online augmentation adaptation.

In summary,
(1) We propose MetaTPT, a novel meta-learning framework that jointly adapts learnable prompts and a self-supervised auxiliary task at test time, thereby enhancing zero-shot generalization of vision-language models.
(2) We introduce a self-supervised task to dynamically optimize augmentations, parameterized as affine transformations, enabling the capture of fine-grained variability within target domains.
(3) We develop consistency-regularized prompt tuning that fosters stable and robust adaptation by enforcing alignment of representations across the learned views.

\section{Preliminaries}
\label{sec:preliminaries}

\paragraph{Prompt tuning for VLMs.}
CLIP~\cite{radford2021learning} consists of two main components: a text encoder $g$ and an image encoder $f$.
Let $\mathbf{t}_y$ denote a handcrafted prompt, such as \textit{``A photo of a \{class\}"}, and let $\mathbf{x}$ represent an image.
The probability of class $y$ given the image $\mathbf{x}$ is:
\begin{equation}
P(\mathbf{x})=\frac{\exp\left(\mathrm{sim}\left(f\left(\mathbf{x}\right), g(\mathbf{t}_y)\right)/\tau\right)}{\sum^{N_c}_{i=1}\exp\left(\mathrm{sim}\left(f\left(\mathbf{x}\right), g(\mathbf{t}_i)\right)/\tau\right)}\ .
\label{eq:clip}
\end{equation}
where $\mathrm{sim}(a,b)=\frac{a^\top b}{\|a\|\|b\|}$ represents the cosine similarity, $\tau$ denotes the temperature, and $N_c$ is the class number.
To adapt CLIP to downstream tasks, prompt-learning methods introduce learnable parameters that modify the input embeddings of the frozen encoders. In CoOp~\cite{zhou2022learning}, a set of learnable text prompt embeddings $\mathbf{\Theta} = \{\mathbf{\Theta}^1, \mathbf{\Theta}^2, \cdots, \mathbf{\Theta}^n\}$ replaces the handcrafted context. MaPLe~\cite{khattak2023maple} extends this paradigm by incorporating learnable prompts $\mathbf{\Theta} = \{\mathbf{\Theta}_{\mathrm{txt}}, \mathbf{\Theta}_{\mathrm{vis}}\}$ for both modalities, where $\mathbf{\Theta}_{\mathrm{txt}} = \{ \mathbf{\Theta}_{\mathrm{txt}}^{(l)} \}^L_{l=1}$ and $\mathbf{\Theta}_{\mathrm{vis}} = \{ \mathbf{\Theta}_{\mathrm{vis}}^{(l)} \}^L_{l=1}$ are inserted across multiple transformer layers, and cross-modal alignment is enforced via a linear projection. MMRL~\cite{guo2025mmrl} further generalizes this framework by inserting learnable prompts  in the higher layers of both encoders, with $\mathbf{\Theta}_{\mathrm{txt}} = \{ \mathbf{\Theta}_{\mathrm{txt}}^{(l)} \}^L_{l=J}$ and $\mathbf{\Theta}_{\mathrm{vis}} = \{ \mathbf{\Theta}_{\mathrm{vis}}^{(l)} \}^L_{l=J}$, allowing independent optimization while maintaining alignment through the contrastive objective.

\paragraph{Test-time prompt tuning.}
TPT~\cite{shu2022test} is a TTA method that optimizes the the learnable prompts $\mathbf{\Theta}$ during inference. For each test sample, it employs fix data augmentation~\cite{hendrycks2019augmix} $ \mathbf{\Phi}$ to generate $N$ augmented views and selects top-$\rho$ confident views for optimization:
\begin{gather}
\min\limits_{\mathbf{\Theta}} H\left(\tilde{P}_{\mathbf{\Theta}}\left(\mathbf{\Phi}(\mathbf{x})\right)\right), \\
\textrm{where}\quad\tilde{P}_{\mathbf{\Theta}}\left(\mathbf{\Phi}(\mathbf{x})\right) = \frac{1}{\rho N}\sum_{i=1}^N \mathds{1}\big[H(P_i)\leqslant\delta\big]P_{\mathbf{\Theta}}\left(\mathbf{\Phi}^i(\mathbf{x})\right).
\label{eq: selection}
\end{gather}
Here, $H(P)=\sum_{i=1}^{N_c}P_i\log P_i$. $\mathbf{\Phi}^i(\mathbf{x})$ denotes the $i$-th augmented view of test sample $\mathbf{x}$, and $\delta$ is the entropy threshold at the $\rho$-th confidence percentile. The learnable prompts $\mathbf{\Theta}$ are optimized by minimizing the average entropy of the selected augmented views.

\paragraph{Meta-learning.}
MAML (Model Agnostic Meta Learning)~\cite{finn2017model} learns a model initialization $\mathbf{\Theta}$ that can quickly adapt to new tasks through a dual-loop optimization:
\begin{equation}
    \min_{\mathbf{\Theta}} \ \mathbb{E}_{\mathcal{T}_i \sim p(\mathcal{T})} \big[ \mathcal{L}_{\mathcal{T}_i} (\mathbf{\Phi}_i) \big],
\label{eq:meta_outer}
\end{equation}
\begin{equation}
    \text{s.t.} \ \mathbf{\Phi}_i = \arg\min_{\mathbf{\Phi}}\mathcal{L}_{\mathcal{T}_i}(\mathbf{\Phi}).
\label{eq:meta_inner}
\end{equation}
Here, Eq.~(\ref{eq:meta_inner}) represents the \textit{inner\ loop}, corresponding to task-specific adaptation, where $\mathbf{\Phi}_i$ is optimized for task $\mathcal{T}_i$. Eq.~(\ref{eq:meta_outer}) defines the \textit{outer\ loop}, updating the meta-parameters~$\mathbf{\Theta}$ to minimize the loss across tasks after adaptation.
\textit{Sun\ et\ al.}~\cite{sun2024learning} extend this framework to the test-time regime. By modeling the RNN hidden state itself with a learnable model parameterized by low-rank matrices $\mathbf{\Phi}_K$ and $\mathbf{\Phi}_V$, which serve as an \textit{inner-loop} learner updated dynamically with incoming inputs. The \textit{outer\ loop} then optimizes the model parameters $\mathbf{\Theta}$ so that the inner-loop updates lead to effective sequence modeling.

\section{Method}
\label{sec:method}

In this section, we introduce \textbf{Meta} \textbf{T}est-time \textbf{P}rompt \textbf{T}uning (\textbf{MetaTPT}), a dual-loop meta-learning framework for adapting vision-language models.
The \textit{inner\ loop} (Sec.~\ref{sec: inner}) learns a self-supervised auxiliary task to optimize parameterized augmentations generating confident and semantically consistent views.
The \textit{outer\ loop} (Sec.~\ref{sec: outer}) updates the learnable prompts by enforcing prediction- and feature-level consistency across these views, enhancing robustness and generalization under domain shifts.

\subsection{Learning a self-supervised task}
\label{sec: inner}
Inspired by TTT~\cite{sun2020test}, we introduce a self-supervised auxiliary task for test-time prompt tuning (TPT)~\cite{shu2022test} to improve its adaptability on unseen target domain. Although TPT effectively adjusts frozen models through prompt optimization, its reliance on fixed augmentation~\cite{hendrycks2019augmix} limits the effectiveness in capturing discriminative features.
To mitigate it, we formulate the self-supervised task as a learnable augmentation strategy. Rather than applying pre-defined transformations, we parameterize the augmentation as an affine transformation $\Phi(\mathbf{x})$:
\begin{equation}
\begin{bmatrix} x' \\ y' \end{bmatrix}=\Phi \begin{bmatrix} x\\ y \\1 \end{bmatrix},\quad\textrm{where}\ \ \Phi=\begin{bmatrix}a & b &t_x\\c & d & t_y\end{bmatrix}.
\end{equation}
where $(x,y)$ is the original coordinates of image $\mathbf{x}$, and $(x', y')$ is the transformed coordinate. 
The affine matrix $\Phi\in\mathbb{R}^{2\times3}$ allows the model to optimize spatial transformations, such as flipping, scaling, aspect ratio adjustments (controlled by $a$ and $d$), shearing (controlled by $b$ and $c$), and translation (controlled by $t_x$ and $t_y$) in a differentiable manner, producing augmented views that are most beneficial for prompt adaptation on each test sample.
Following TPT~\cite{shu2022test}, we generate a batch of $N$ augmented views per sample to introduce diversity, denoting the set of affine matrices as $\mathbf{\Phi} = \{\Phi^i\}_{i=1}^N\in\mathbb{R}^{N\times2\times3}$.

To ensure the augmented views effectively guide subsequent prompt adaptation, $\mathbf{\Phi}$ are optimized to satisfy two objectives.
First, they should generate augmented views with high confidence by minimizing the prediction entropy:
\begin{gather}
     H(\mathbf{x};\mathbf{\Phi};\mathbf{\Theta})=-\sum_{i=1}^{N_c}\hat{P}^i_{\mathbf{\Theta}}(\mathbf{\Phi}(\mathbf{x})) \log\hat{P}^i_{\mathbf{\Theta}}(\mathbf{\Phi}(\mathbf{x})),\\
     \textrm{where}\quad\hat{P}_{\mathbf{\Theta}}(\mathbf{\Phi}(\mathbf{x}))=P_{\mathbf{\Theta}}(\mathbf{x})+\tilde{P}_{\mathbf{\Theta}}(\mathbf{\Phi}(\mathbf{x})).
     \label{eq: entropy}
\end{gather}
Here, the composite entropy over both the input $\mathbf{x}$ and its augmented views $\mathbf{\Phi}(\mathbf{x})$ empirically outperforms optimizing the views alone.
Second, the augmentations should avoid semantic bias by preserving the intrinsic content of the original image. This is enforced by minimizing the feature-level discrepancy between the original input $\mathbf{x}$ and the averaged features of its augmented views $\mathbf{\Phi}(\mathbf{x})$:
\begin{gather}
    \mathcal{L}_{\text{dis}}(\mathbf{x};\mathbf{\Phi}; \mathbf{\Theta}) = \left\| f_{\mathbf{\Theta}}(\mathbf{x}) - \bar{f}_{\mathbf{\Theta}}(\mathbf{\Phi}(\mathbf{x})) \right\|_2 ,\\
 \textrm{where}\quad\bar{f}(\mathbf{\Phi}_{\mathbf{\Theta}}(\mathbf{x})) = \frac{1}{N} \sum_{i=1}^N f_{\mathbf{\Theta}}(\mathbf{\Phi}^i(\mathbf{x})).
\label{eq: feature_in}
\end{gather}
These two items are combined to optimize $\mathbf{\Phi}$ in the inner loop:
\begin{equation}
\mathcal{L}_\textrm{inner} =  H(\mathbf{x};\mathbf{\Phi};\mathbf{\Theta}) + \mathcal{L}_{\text{dis}}(\mathbf{x};\mathbf{\Phi};\mathbf{\Theta}).
\label{eq: loss_in}
\end{equation}

We introduce two sets of affine matrices, 
$\mathbf{\Phi}_K$ and $\mathbf{\Phi}_V$, each initialized with distinct augmentation types to generate diverse input patterns for prompt adaptation.
Directly optimizing may lead them to converge toward similar transformation patterns, reducing augmentation diversity.
To prevent this, we optimize $\mathbf{\Phi}_K$ via Eq.~(\ref{eq: loss_in}), while $\mathbf{\Phi}_V$ is updated using an Exponential Moving Average (EMA)~\cite{holt2004forecasting} of $\mathbf{\Phi}_K$:
\begin{equation}
\mathbf{\Phi}_V \gets \alpha \mathbf{\Phi}_V + (1 - \alpha) \mathbf{\Phi}_K.
\label{loss: ema}
\end{equation}
where $\alpha$ is the momentum coefficient. Note that in the inner loop, only the affine matrices $\mathbf{\Phi}_K$ and $\mathbf{\Phi}_V$ are updated, while the learnable prompts $\mathbf{\Theta}$ remain frozen.

\subsection{Consistency-regularized prompt tuning}
\label{sec: outer}
Entropy-based objectives, as employed in TPT, can promote confident predictions; however, under challenging domain shifts, they often produce overconfident yet inaccurate outputs.
To mitigate this, MetaTPT enforces consistency between two sets of learned views, $\mathbf{\Phi}_K(\mathbf{x})$ and $\mathbf{\Phi}_V(\mathbf{x})$, providing a reliable supervisory signal for optimizing the learnable prompts $\mathbf{\Theta}$.

Predictive consistency $\mathcal{L}_{ce}$ is first enforced in the probability space through a cross-entropy loss between the soft predictions from two distinct sets of augmented views:
\begin{equation}
     \mathcal{L}_{ce}(\mathbf{x}; \mathbf{\Phi}_K;\mathbf{\Phi}_V;\mathbf{\Theta}) = -\sum_{i=1}^{N_c}\hat{P}^i_{\mathbf{\Theta}}(\mathbf{\Phi}_K(\mathbf{x})) \log\hat{P}^i_{\mathbf{\Theta}}(\mathbf{\Phi}_V(\mathbf{x})).
\label{eq: cross_entropy}
\end{equation}
Semantic consistency $\mathcal{L}_\textrm{dis}$ is further enforced in the feature space by minimizing the distance between the average feature representations of the two augmented views:
\begin{equation}
    \mathcal{L}_\textrm{dis}(\mathbf{x};\mathbf{\Theta}_K; \mathbf{\Theta}_V; \mathbf{R}) = \| \bar{f}_{\mathbf{\Theta}}(\mathbf{\Phi}_K(\mathbf{x})) - \bar{f}_{\mathbf{\Theta}}(\mathbf{\Phi}_V(\mathbf{x})) \|_2.
\label{eq: feature_ou}
\end{equation}
The learnable prompts $\mathbf{\Theta}$ are jointly optimized by two objectives, while $\mathbf{\Phi}_K$ and $\mathbf{\Phi}_V$ remain fixed:
\begin{equation}
    \mathcal{L}_\textrm{outer} =  \mathcal{L}_{ce}(\mathbf{x}; \mathbf{\Phi}_K;\mathbf{\Phi}_V;\mathbf{\Theta}) + \mathcal{L}_\textrm{dis}(\mathbf{x};\mathbf{\Phi}_K; \mathbf{\Phi}_V; \mathbf{\Theta}).
\label{eq: loss_ou}
\end{equation}

\subsection{Dual-loop meta adaptation}

\begin{wrapfigure}{r}{0.6\linewidth}
\vspace{-22pt}
\begin{minipage}{\linewidth}
\begin{algorithm}[H]
\caption{MetaTPT: Meta Test-Time Prompt Tuning}
\label{alg:metatpt}
\begin{algorithmic}[1]
\REQUIRE Target distribution $p(\mathcal{X})$; learnable prompts $\mathbf{\Theta}$; batch size $N$; loop steps $T, M$; learning rate $\eta_i, \eta_o$; EMA momentum $\alpha$; weights $\lambda_K, \lambda_V$.
\FOR{all test sample $\mathbf{x}\sim p(\mathcal{X}))$}
    \STATE Initialize $\mathbf{\Theta}$ from a ImageNet-pretrained model.
    \STATE Initialize $\mathbf{\Phi}_K^{i\in[1,N]}$ using Eq.~(\ref{eq: phi_K}), and $\mathbf{\Phi}_V^{i\in[1,N]}$ using Eq.~(\ref{eq: phi_V}).
    \FOR{$m = 1$ to $M$}
        \FOR{$t = 1$ to $T$}
            \STATE \textcolor{blue}{// Inner loop: learning a self-supervised task}
            \STATE $\mathcal{L}_{\text{inner}} = H(\mathbf{x}; \mathbf{\Phi}_K; \mathbf{\Theta}) + \mathcal{L}_{\text{dis}}(\mathbf{x};\mathbf{\Phi}_K; \mathbf{\Theta})$.
            \STATE Update $\mathbf{\Phi}_K \leftarrow \mathbf{\Phi}_K - \eta_i \nabla_{\mathbf{\Phi}_K} \mathcal{L}_{\text{inner}}$. 
            \STATE Update $\mathbf{\Phi}_V \leftarrow \alpha \mathbf{\Phi}_V + (1 - \alpha) \mathbf{\Phi}_K$.
        \ENDFOR
            \STATE \textcolor{blue}{// Outer loop: consistency-regularized prompt tuning}
            \STATE $\mathcal{L}_{\text{outer}} = \mathcal{L}_{ce}(\mathbf{x};\mathbf{\Phi}; \mathbf{\Theta}) + \mathcal{L}_{\text{dis}}(\mathbf{x};\mathbf{\Phi}; \mathbf{\Theta})$.
            \STATE Update $\mathbf{\Theta} \leftarrow \mathbf{\Theta} - \eta_o \nabla_{\mathbf{\Theta}} \mathcal{L}_{\text{outer}}$.
    \ENDFOR
    \STATE $\hat{P}(\mathbf{x})=P_{\mathbf{\Theta}}(\mathbf{x})+\lambda_K \tilde{P}_{\mathbf{\Theta}}(\mathbf{\Phi}_K(\mathbf{x}))+\lambda_V \tilde{P}_{\mathbf{\Theta}}(\mathbf{\Phi}_V(\mathbf{x}))$. 
\ENDFOR
\end{algorithmic}  
\end{algorithm}
\end{minipage}
\end{wrapfigure}

For each test sample $\mathbf{x}$ from the target domain, the model is adapted using a dual-loop meta-learning framework:
\begin{equation}
    \min_{\mathbf{\Theta}} \ \mathbb{E}_{\mathbf{x} \sim p(\mathcal{X})} \big[ \mathcal{L}_{\textrm{outer}} (\mathbf{x};\mathbf{\Phi};\mathbf{\Theta}) \big],
    \label{eq:metatpt_outer}
\end{equation}
\begin{equation}
    \text{s.t.} \ \mathbf{\Phi} = \arg\min_{\mathbf{\Phi}}\mathcal{L}_{\textrm{inner}}(\mathbf{x};\mathbf{\Phi};\mathbf{\Theta}).
    \label{eq:metatpt_inner}
\end{equation}
Here, Eq.~(\ref{eq:metatpt_inner}) defines the \textit{inner loop}, which minimizes Eq.~(\ref{eq: loss_in}) to learn a self-supervised task, thereby optimizing the learnable augmentations $\mathbf{\Phi}$ to generate informative views for sample $\mathbf{x}$.
Eq.~(\ref{eq:metatpt_outer}) represents the \textit{outer loop}, which updates the learnable prompts $\mathbf{\Theta}$ by minimizing Eq.~(\ref{eq: loss_ou}).
This framework enables sample-wise adaptation of the test stream, enhancing zero-shot generalization.

After adaptation, the final prediction is obtained by aggregating information from the original input $\mathbf{x}$ and its selected augmented views.
\begin{equation}
    \hat{P}_{\mathbf{\Theta}}(\mathbf{x})=P_{\mathbf{\Theta}}(\mathbf{x})+\lambda_K \tilde{P}_{\mathbf{\Theta}}(\mathbf{\Phi}_K(\mathbf{x}))+\lambda_V \tilde{P}_{\mathbf{\Theta}}(\mathbf{\Phi}_V(\mathbf{x})).
\end{equation}
where $\lambda_K$ and $\lambda_V$ weight the contributions of the augmented views.

\section{Experiments}
\label{sec:experiment}

\subsection{Experimental Setup}


\paragraph{Implementation details.}
Following the test-time adaptation (TTA) protocol, we adapt the source-trained model to the target domain data during inference. Following TPT~\cite{shu2022test}, we use ImageNet as the target domain and fine-tune the learnable prompts in a few-shot setting for three vision-language models: CoOp, MaPLe, and MMRL. We then evaluate MetaTPT on these models and the zero-shot CLIP across target datasets. Unless otherwise specified, all models employed CLIP-ViT-B/16 as the visual backbone with default hyperparameters. In Alg.~\ref{alg:metatpt}, we set $M=T=1$, adopting a single inner and outer optimization loop for a fair comparison. Both loops use the AdamW optimizer, with learning rates $\eta_i=\eta_o=0.0001$ for domain generation and $0.001$ for cross-dataset evaluation. The EMA momentum was set to $\alpha=0.9$, and a grid search is performed over $\lambda_K$ and $\lambda_V$.
Following TPT~\cite{shu2022test}, we generate $N=64$ augmented views for both $\mathbf{\Phi}_K$ and $\mathbf{\Phi}_V$, each with different initializations. Specifically, $\mathbf{\Phi}_V$ is initialized to mimic the \textit{rotation} task used in TTT~\cite{sun2020test}:
\begin{equation}
\mathbf{\Phi}_V^{i} =
\begin{bmatrix}
\cos(\gamma) & -\sin(\gamma) & 0 \\ 
\sin(\gamma) & \cos(\gamma) & 0
\end{bmatrix},
\ i=\{1,\cdots,N\}.
\label{eq: phi_V}
\end{equation}
Each $\mathbf{\Phi}_V^{i\in[1, 64]}$ is initialized with Eq.~(\ref{eq: phi_V}), with a rotation angle $\gamma$ sampled uniformly from $(0^\circ, 30^\circ)$, enabling in-plane rotations of up to $30$ degrees.
$\mathbf{\Phi}_K$ is initialized to mimic the \textit{Random Resize Crop} and \textit{Random Horizontal Flip} augmentations from PromptAlign~\cite{abdul2023align}:
\begin{equation}
\mathbf{\Phi}_K^{i} =
\begin{bmatrix}
\frac{\mathrm{flip}\cdot w}{\mathrm{width}} & 0 & t_x \\ 
0 & \frac{h}{\mathrm{height}} & t_y
\end{bmatrix},
\ \ i=\{1\cdots N\}.
\label{eq: phi_K}
\end{equation}
where $\mathrm{flip} \in \{-1, 1\}$  is a random $x$-axis flipping factor, emulating \textit{Random Horizontal Flip}.
$\mathrm{width}$ and $\mathrm{height}$ are the original dimensions of image $\mathbf{x}$.
$w$ and $h$ represent the new dimensions: $w= \sqrt{\mathrm{targetArea} \cdot\rm{ratio}},\ h = \sqrt{\frac{\mathrm{targetArea}}{\rm{ratio}}}$, where $\mathrm{targetArea} = \mathrm{scale}\cdot \mathrm{width}\cdot \mathrm{height}.$ Here, the hyperparameters $\mathrm{scale}$ and $\mathrm{ratio}$ mimic the input parameters of \textit{Random Resize Crop}.
If valid dimensions  $w$ and $h$ are found, random starting coordinates $(i, j)$ are selected to prevent the image from being cropped beyond its boundaries: $i\in\left[0,\mathrm{height}-h\right],\ j\in\left[0,\mathrm{width}-w\right].$
$t_x$ and $t_y$ translate the center of the original image $\left(\frac{\mathrm{width}}{2}, \frac{\mathrm{height}}{2}\right)$ to the new center $\left(j+\frac{w}{2},i+\frac{h}{2}\right)$.
Each $\mathbf{\Phi}_K^{i \in [1, 64]}$ is initialized via Eq.~(\ref{eq: phi_K}), where $\mathrm{scale}$ and $\mathrm{ratio}$ are randomly sampled. Specifically, $\mathrm{scale}$ is drawn from the interval $(0.2, 1.0)$, resizing the image to $20\%\sim100\%$ of its original area. The aspect $\mathrm{ratio}$ is sampled from $(\frac{3}{4}, \frac{4}{3})$, introducing mild geometric distortions.
All initialization hyperparameters are fixed and shared across target datasets. Notably, during inner-loop optimization, we update the affine matrices $\mathbf{\Phi}_K$ and $\mathbf{\Phi}_V$ directly, rather than tuning their initialization parameters. These hyperparameters are used only to construct the initial affine matrices.
We implement MetaTPT on a NVIDIA A800 80GB GPU. For CoOp, MaPLe and MMRL, all results are averaged over three independent runs.

\subsection{Results}
\paragraph{Domain generalization.}
Table~\ref{tab:domain-generation} reports the performance of four vision-language models across four out-of-distribution ImageNet variants. The results demonstrate that our MetaTPT outperforms other adaptation methods across all base models. By contrast, TPT exhibits limited robustness and may underperform under certain conditions; for example, for MMRL, TPT slightly decreases performance on ImageNet-R by -0.22\%, whereas MetaTPT improves it by +2.78\%. Similar gains are observed on ImageNet-A and ImageNet-Sketch, indicating that MetaTPT effectively mitigates performance degradation and enhances adaptation to diverse target domains.

\paragraph{Cross-dataset evaluation.}
Table~\ref{tab:cross-dataset-evaluation} presents the performance of various vision-language models across ten image classification datasets. The results underscore the superior adaptability of our MetaTPT, particularly in challenging domain shifts. For example, while TPT fails to effectively adapt MaPLe on DTD (-0.62\%) and EuroSAT (-0.26\%), MetaTPT achieves substantial gains of +1.86\% and +1.43\%, respectively. Furthermore, TPT exhibits limited adaptability for MMRL, whereas MetaTPT consistently improves performance across all datasets, thereby demonstrating its efficacy in cross-domain adaptation.

\begin{table}[t]
\centering
\tabcolsep=0.4cm
\caption{Comparison of MetaTPT on \textbf{Domain Generation} is conducted across four variants.}
\label{tab:domain-generation}
\resizebox{\textwidth}{!}{%
\begin{tabular}{l|cccc|c}
\toprule
 & ImageNet-V2 & ImageNet-Sketch & ImageNet-A & ImageNet-R & \textit{Average} \\
 \midrule
CLIP~\cite{radford2021learning} & 60.86 & 46.09 & 47.87 & 73.98 & 57.20 \\
CLIP + TPT~\cite{shu2022test} & 63.45 & 47.94 & 54.77 & 77.06 & 60.81 \\
\rowcolor[HTML]{E1E1E1} \textbf{CLIP + MetaTPT} & 63.87 & 47.97 & 60.04 & 76.73 & 62.15 \\
\midrule
CoOp~\cite{zhou2022learning} & 64.20 & 47.99 & 49.71 & 75.21 & 59.28 \\
CoOp + TPT~\cite{shu2022test} & 66.83 & 49.29 & 57.95 & 77.27 & 62.84 \\
\rowcolor[HTML]{E1E1E1} \textbf{CoOp + MetaTPT} & \textbf{67.03} & 49.36 & 62.80 & 77.78 & 64.24 \\
\midrule
MaPLe~\cite{khattak2023maple} & 64.07 & 49.15 & 50.90 & 76.98 & 60.28 \\
MaPLe + TPT~\cite{shu2022test} & 64.87 & 48.16 & 58.08 & 78.12 & 62.31 \\
MaPLe + PromptAlign~\cite{abdul2023align} & 65.29 & 50.23 & 59.37 & 79.33 & 63.56 \\
\rowcolor[HTML]{E1E1E1} 
\textbf{MaPLe + MetaTPT} & 66.46 & 51.37 & \textbf{62.83} & 79.63 & \textbf{65.07} \\
\midrule
MMRL~\cite{guo2025mmrl} & 64.47 & 49.17 & 51.20 & 77.53 & 60.59 \\
MMRL + TPT & 64.49 & 49.09 & 50.26 & 77.31 & 60.29 \\
\rowcolor[HTML]{E1E1E1} 
\textbf{MMRL + MetaTPT} & 66.40 & \textbf{51.49} & 58.47 & \textbf{80.31} & 64.17 \\
\bottomrule
\end{tabular}
} %
\end{table}

\begin{table}[t]
\centering
\caption{Comparison of MetaTPT on \textbf{Cross-Dataset Evaluation} is conducted across ten datasets.}
\label{tab:cross-dataset-evaluation}
\resizebox{\textwidth}{!}{%
\begin{tabular}{l|cccccccccc|c}
\toprule
& \rotatebox{90}{Caltech101} & \rotatebox{90}{OxfordPets} & \rotatebox{90}{StanfordCars} & \rotatebox{90}{OxfordFlowers} & \rotatebox{90}{Food101} & \rotatebox{90}{FGVCAircraft} & \rotatebox{90}{SUN397} & \rotatebox{90}{DTD} & \rotatebox{90}{EuroSAT} & \rotatebox{90}{UCF101} & \rotatebox{90}{\textit{Average}}  \\ \midrule
CLIP~\cite{radford2021learning} & 93.35 & 88.25 & 65.48 & 67.44 & 83.65 & 23.67 & 62.59 & 44.27 & 42.01 & 65.13 & 63.58 \\
CLIP + TPT~\cite{shu2022test} & 94.16 & 87.79 & 66.87 & 68.98 & 84.67 & 24.78 & 65.50 & 47.75 & 42.44 & 68.04 & 65.12 \\
\rowcolor[HTML]{E1E1E1} \textbf{CLIP + MetaTPT} & 94.81 & 90.57 & 68.71 & 70.77 & 86.71 & 26.52 & 66.45 & 48.17 & 41.96 & 69.92 & 66.46 \\
\midrule
CoOp~\cite{zhou2022learning} & 93.70 & 89.14 & 64.51 & 68.71 & 85.30 & 18.47 & 64.15 & 41.92 & 46.39 & 66.55 & 63.88 \\
CoOp + TPT~\cite{shu2022test} & 93.15 & 89.48 & 66.77 & 68.48 & 86.48 & 20.51 & 66.06 & 43.32 & 37.73 & 68.91 & 64.09 \\
\rowcolor[HTML]{E1E1E1} \textbf{CoOp + MetaTPT} & 93.59 & 90.76 & 67.24 & 69.14 & 86.81 & 21.97 & 66.77 & 45.37 & 41.74 & 70.43 & 65.38 \\
\midrule
MaPLe~\cite{khattak2023maple} & 93.53 & 90.46 & 65.57 & 72.23 & 86.20 & 24.74 & 67.01 & 46.49 & 48.06 & 68.69 & 66.30 \\
MaPLe + TPT~\cite{shu2022test} & 93.59 & 90.72 & 66.50 & 72.37 & 86.64 & 24.70 & 67.54 & 45.87 & 47.80 & 69.19 & 66.49 \\
MaPLe + PromptAlign~\cite{abdul2023align} & 94.01 & 90.76 & 68.50 & 72.39 & 86.65 & 24.80 & 67.54 & 47.24 & 47.86 & 69.47 & 66.92 \\
\rowcolor[HTML]{E1E1E1} 
\textbf{MaPLe + MetaTPT} & 94.31 & 90.82 & 68.69 & 72.70 & 87.28 & 26.41 & 68.16 & 48.35 & 49.49 & 69.87 & 67.61  \\
\midrule
MMRL~\cite{guo2025mmrl} & 94.67 & 91.43 & 66.10 & 72.77 & 86.40 & 26.30 & 67.57 & 45.90 & 53.10 & 68.27 & 67.25 \\
MMRL + TPT & 94.40 & 91.31 & 66.49 & 72.89 & 86.18 & 26.23 & 67.27 & 46.41 & 47.18 & 69.04 & 66.74 \\
\rowcolor[HTML]{E1E1E1} 
\textbf{MMRL + MetaTPT} & \textbf{94.90} & \textbf{92.79} & \textbf{69.50} & \textbf{74.22} & \textbf{87.61} & \textbf{29.05} & \textbf{69.17} & \textbf{48.88} & \textbf{54.26} & \textbf{72.24} & \textbf{69.26} \\
\bottomrule
\end{tabular}%
} %
\end{table}

\begin{figure}[t]
    \centering
    \includegraphics[width=\linewidth]{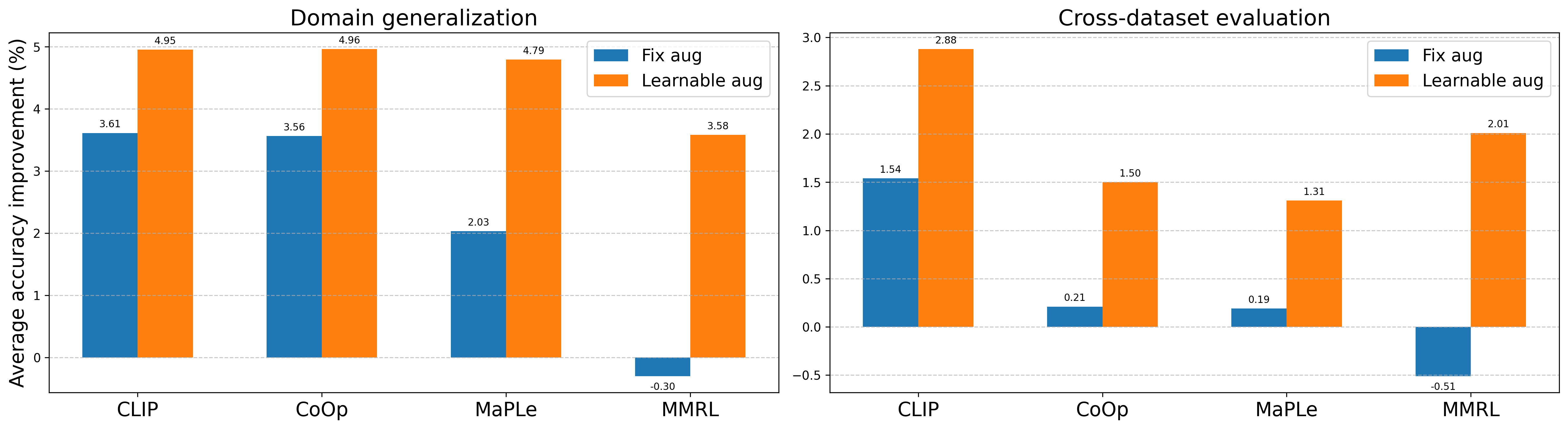}
    \caption{\textbf{Effect of learnable augmentation}: comparison of the average performance gains achieved by learnable augmentations in MetaTPT versus fixed augmentations in TPT across the base models.}
    \label{fig:ab_aug}
\end{figure}

\begin{figure}[t]
  \centering
  \begin{minipage}[t]{0.48\textwidth}
    \centering
    \includegraphics[width=\textwidth]{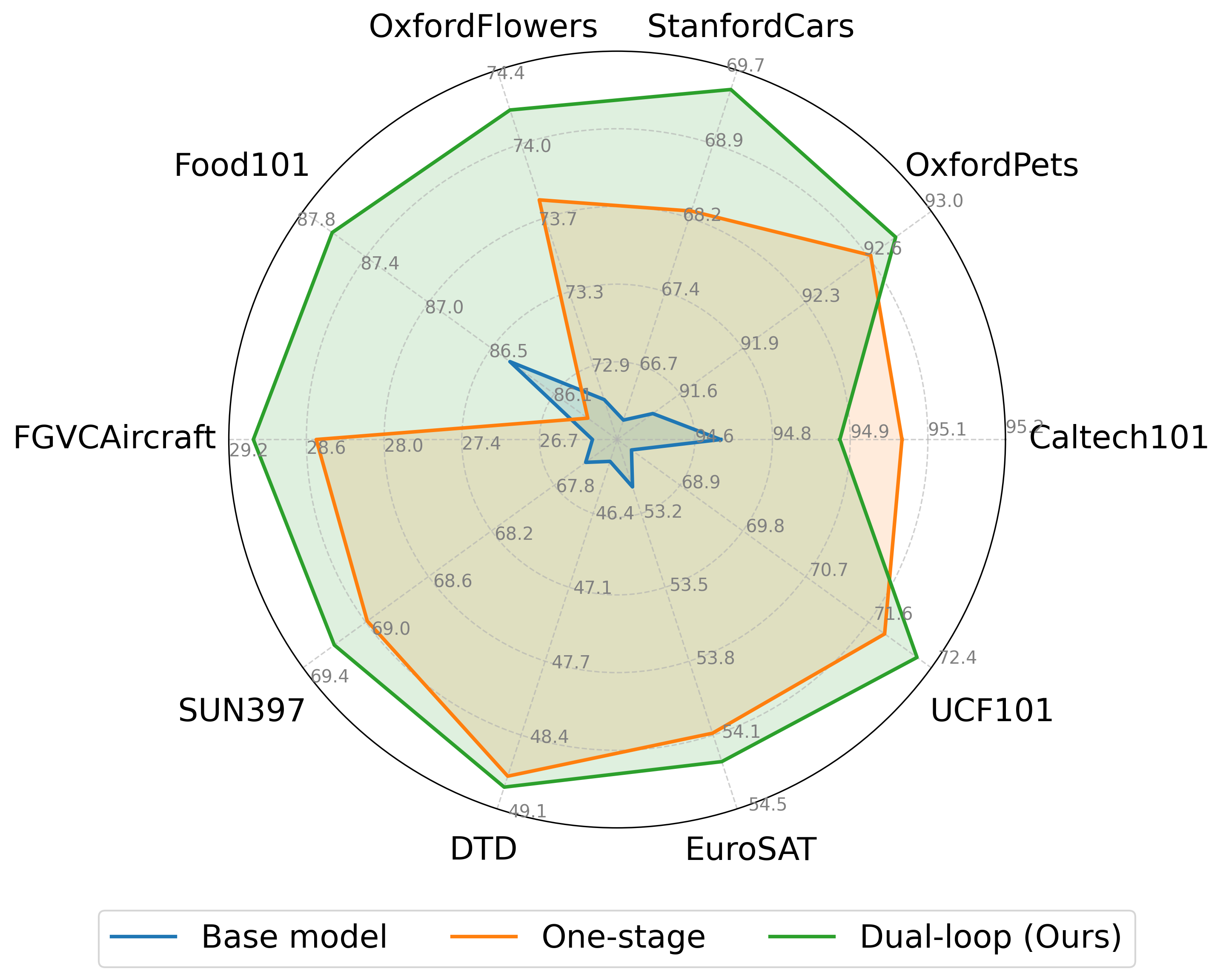}
    \caption{\textbf{Effect of dual-loop optimization}: the “one-stage” scheme updates $\mathbf{\Phi}$ and $\mathbf{\Theta}$ simultaneously, whereas our “dual-loop” scheme alternates their updates in an interleaved manner.}
    \label{fig:ab_dual}
  \end{minipage}
  \hfill
  \begin{minipage}[t]{0.48\textwidth}
    \centering
    \includegraphics[width=\textwidth]{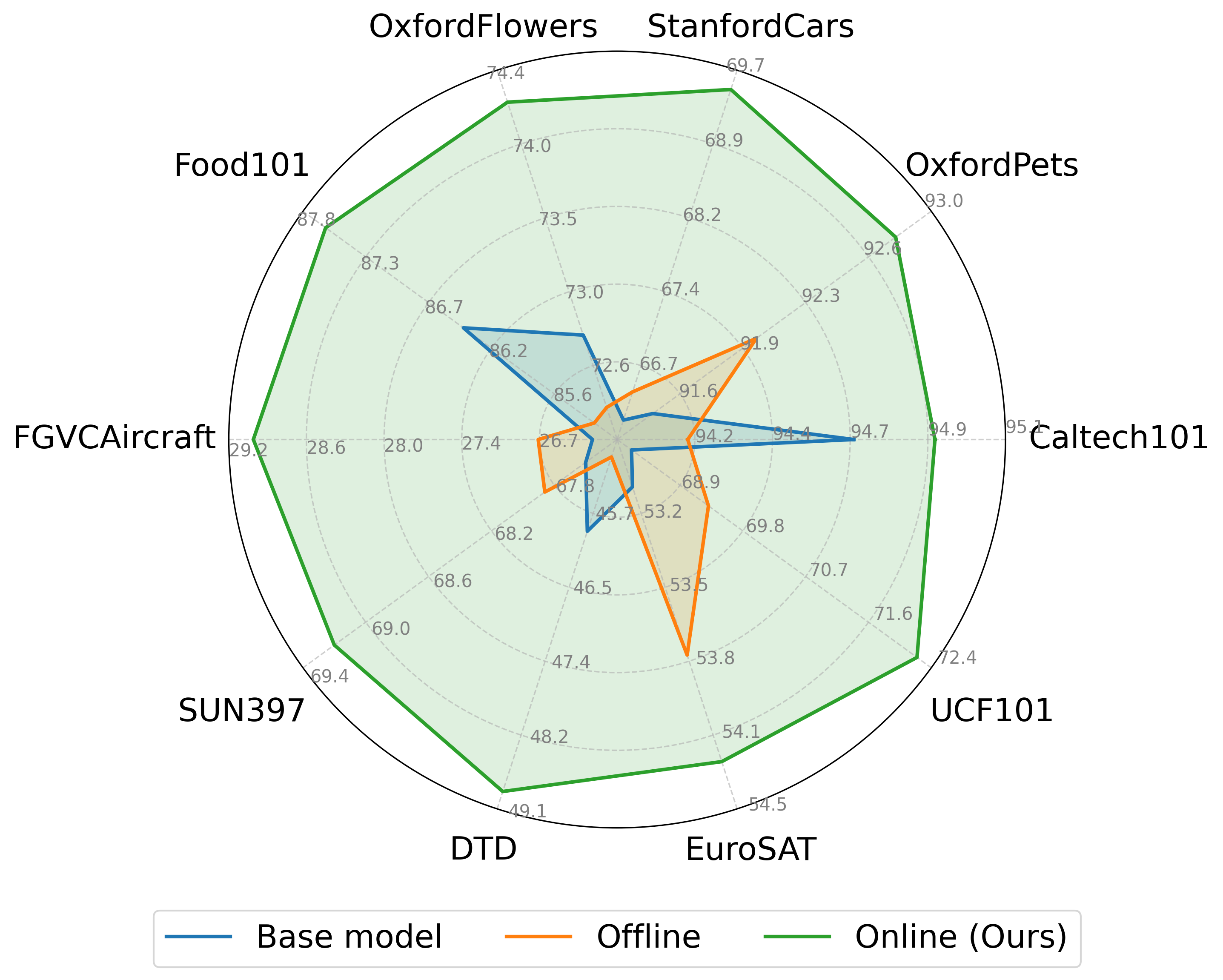}
    \caption{\textbf{Effect of online adaptation}: In the “offine” setting, $\mathbf{\Phi}$ is meta-trained prior to prompt adaptation, whereas in our “online” setting, $\mathbf{\Phi}$ and $\mathbf{\Theta}$ are meta-adapted per test sample.}
    \label{fig:ab_online}
  \end{minipage}
\end{figure}

\begin{figure}[t]
    \centering
    \includegraphics[width=0.7\linewidth]{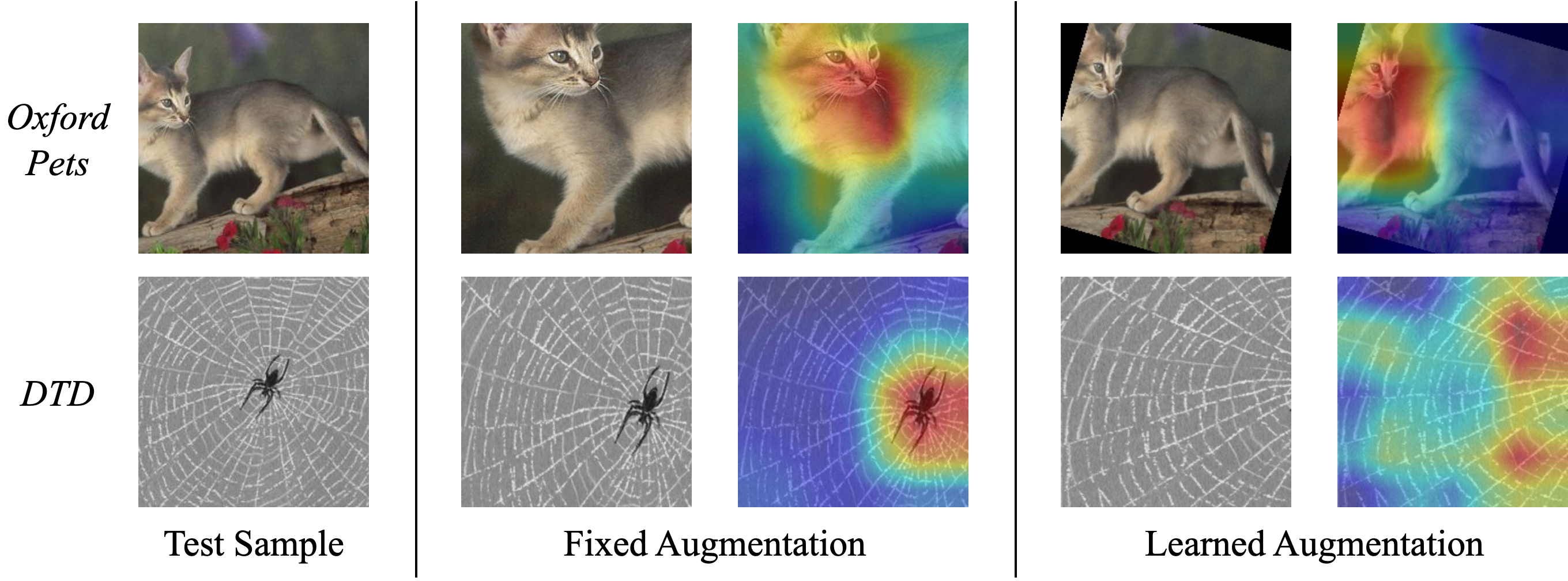}
    \caption{\textbf{Visualization of learnable augmentations} and corresponding attention maps.}
    \label{fig:visualization}
\end{figure}

\subsection{Ablation Studies}
\paragraph{Effect of learnable augmentations.}
Figure~\ref{fig:ab_aug} presents the effectiveness of learnable augmentations within MetaTPT for adapting diverse vision-language models. The base model is trained on the source domain and evaluated directly on the target domain, providing a baseline for measuring adaptation performance. To mitigate domain shift, TPT employs fixed augmentations for test-time prompt tuning, which can improve target-domain performance in many cases; however, its benefits are not universal. For instance, TPT yields negative gains on MMRL, with -0.30\% in domain generalization and -0.51\% in cross-dataset evaluation. In contrast, MetaTPT introduces learnable augmentations that are dynamically optimized at test time, consistently improving performance across all base models in both cross-dataset evaluation and domain generalization.

\paragraph{Effect of dual-loop optimization.}
Figure~\ref{fig:ab_dual} illustrates the effectiveness of dual-loop optimization within MetaTPT. Compared to one-stage training, where the learnable prompts $\mathbf{\Theta}$ and learnable augmentations $\mathbf{\Phi}$ are jointly optimized under a unified objective, $\mathcal{L} = \mathcal{L}_{\text{inner}} + \mathcal{L}_{\text{outer}}$, the dual-loop approach decouples the updates of $\mathbf{\Phi}$ and $\mathbf{\Theta}$ into dedicated inner and outer loops, respectively. While the one-stage optimization is straightforward, it fails to capture the hierarchical dependency between primary and auxiliary tasks. In contrast, the dual-loop design aligns with the meta-learning principle of nested optimization, facilitating more effective task adaptation and generalization. Empirically, dual-loop optimization increases the average accuracy for MMRL from 68.78\% to 69.26\%, demonstrating the benefits of respecting the intrinsic hierarchical structure of meta-learning.

\paragraph{Effect of online augmentation adaptation.}
Figure~\ref{fig:ab_online} demonstrates the effectiveness of online adaptation of learnable augmentations in MetaTPT. Following the online setting of TPT, MetaTPT meta-adapts both prompts $\mathbf{\Theta}$ and augmentations $\mathbf{\Phi}$ on a per-sample basis within a single test stream. To provide a comparative baseline, we introduce an offline variant, in which augmentations $\mathbf{\Phi}$ are first meta-trained for each target domain and subsequently applied to each test sample during meta-testing with prompt tuning. Notably, the offline variant exhibits a performance degradation of 0.39\% relative to the online approach, underscoring the superiority of online adaptive augmentation in enhancing test-time generalization.


\paragraph{Visualization of learnable augmentations.}
Figure~\ref{fig:visualization} visualizes the augmented views and their corresponding attention maps. In natural-object domains such as OxfordPets, fixed augmentations may crop views that still retain informative object content, allowing the model to focus its attention on relevant regions. However, in more challenging domains like DTD, fixed augmentations are more likely to crop out domain-relevant features—for example, producing a view that highlights a spider rather than a cobweb (the ground-truth class). In contrast, our learnable augmentation generates views that better preserve domain-specific semantics, such as emphasizing cobweb textures to guide the model’s attention. Interestingly, in OxfordPets, the learnable augmentation frequently rotates objects (e.g., Abyssinian) to a horizontal orientation,  making it easier for the model to recognize.

\section{Conclusion}
\label{sec:conclusion}

In this work, we proposed MetaTPT, a meta-learning framework for test-time adaptation that facilitates zero-shot generalization.
By jointly optimizing parameterized augmentations and learnable prompts via dual-loop optimization, MetaTPT effectively addresses the limitations of previous methods that rely on fixed augmentation strategies.
Experiments on two benchmarks show that MetaTPT consistently outperforms existing methods under distribution shifts.

\bibliographystyle{abbrv}
\bibliography{ref}

@inproceedings{jia2021scaling,
  title={Scaling up visual and vision-language representation learning with noisy text supervision},
  author={Jia, Chao and Yang, Yinfei and Xia, Ye and Chen, Yi-Ting and Parekh, Zarana and Pham, Hieu and Le, Quoc and Sung, Yun-Hsuan and Li, Zhen and Duerig, Tom},
  booktitle={International conference on machine learning},
  pages={4904--4916},
  year={2021},
  organization={PMLR}
}

@inproceedings{radford2021learning,
  title={Learning transferable visual models from natural language supervision},
  author={Radford, Alec and Kim, Jong Wook and Hallacy, Chris and Ramesh, Aditya and Goh, Gabriel and Agarwal, Sandhini and Sastry, Girish and Askell, Amanda and Mishkin, Pamela and Clark, Jack and others},
  booktitle={International conference on machine learning},
  pages={8748--8763},
  year={2021},
  organization={PmLR}
}

@inproceedings{li2022blip,
  title={Blip: Bootstrapping language-image pre-training for unified vision-language understanding and generation},
  author={Li, Junnan and Li, Dongxu and Xiong, Caiming and Hoi, Steven},
  booktitle={International conference on machine learning},
  pages={12888--12900},
  year={2022},
  organization={PMLR}
}

@article{alayrac2022flamingo,
  title={Flamingo: a visual language model for few-shot learning},
  author={Alayrac, Jean-Baptiste and Donahue, Jeff and Luc, Pauline and Miech, Antoine and Barr, Iain and Hasson, Yana and Lenc, Karel and Mensch, Arthur and Millican, Katherine and Reynolds, Malcolm and others},
  journal={Advances in neural information processing systems},
  volume={35},
  pages={23716--23736},
  year={2022}
}

@article{yu2022coca,
  title={Coca: Contrastive captioners are image-text foundation models},
  author={Yu, Jiahui and Wang, Zirui and Vasudevan, Vijay and Yeung, Legg and Seyedhosseini, Mojtaba and Wu, Yonghui},
  journal={Transactions on Machine Learning Research},
  year={2022}
}

@article{shu2022test,
  title={Test-time prompt tuning for zero-shot generalization in vision-language models},
  author={Shu, Manli and Nie, Weili and Huang, De-An and Yu, Zhiding and Goldstein, Tom and Anandkumar, Anima and Xiao, Chaowei},
  journal={Advances in Neural Information Processing Systems},
  volume={35},
  pages={14274--14289},
  year={2022}
}

@article{hendrycks2019augmix,
  title={Augmix: A simple data processing method to improve robustness and uncertainty},
  author={Hendrycks, Dan and Mu, Norman and Cubuk, Ekin D and Zoph, Barret and Gilmer, Justin and Lakshminarayanan, Balaji},
  journal={arXiv preprint arXiv:1912.02781},
  year={2019}
}

@inproceedings{feng2023diverse,
  title={Diverse data augmentation with diffusions for effective test-time prompt tuning},
  author={Feng, Chun-Mei and Yu, Kai and Liu, Yong and Khan, Salman and Zuo, Wangmeng},
  booktitle={Proceedings of the IEEE/CVF International Conference on Computer Vision},
  pages={2704--2714},
  year={2023}
}

@inproceedings{rombach2022diffusion,
  title={High-resolution image synthesis with latent diffusion models},
  author={Rombach, Robin and Blattmann, Andreas and Lorenz, Dominik and Esser, Patrick and Ommer, Bj{\"o}rn},
  booktitle={Proceedings of the IEEE/CVF conference on computer vision and pattern recognition},
  pages={10684--10695},
  year={2022}
}

@article{abdul2023align,
  title={Align your prompts: Test-time prompting with distribution alignment for zero-shot generalization},
  author={Abdul Samadh, Jameel and Gani, Mohammad Hanan and Hussein, Noor and Khattak, Muhammad Uzair and Naseer, Muhammad Muzammal and Shahbaz Khan, Fahad and Khan, Salman H},
  journal={Advances in Neural Information Processing Systems},
  volume={36},
  pages={80396--80413},
  year={2023}
}

@article{zhang2024historical,
  title={Historical test-time prompt tuning for vision foundation models},
  author={Zhang, Jingyi and Huang, Jiaxing and Zhang, Xiaoqin and Shao, Ling and Lu, Shijian},
  journal={Advances in Neural Information Processing Systems},
  year={2024}
}

@inproceedings{karmanov2024efficient,
  title={Efficient test-time adaptation of vision-language models},
  author={Karmanov, Adilbek and Guan, Dayan and Lu, Shijian and El Saddik, Abdulmotaleb and Xing, Eric},
  booktitle={Proceedings of the IEEE/CVF Conference on Computer Vision and Pattern Recognition},
  pages={14162--14171},
  year={2024}
}

@article{zhang2024boostadapter,
  title={BoostAdapter: Improving Vision-Language Test-Time Adaptation via Regional Bootstrapping},
  author={Zhang, Taolin and Wang, Jinpeng and Guo, Hang and Dai, Tao and Chen, Bin and Xia, Shu-Tao},
  journal={Advances in Neural Information Processing Systems},
  year={2024}
}

@article{zhou2022learning,
  title={Learning to prompt for vision-language models},
  author={Zhou, Kaiyang and Yang, Jingkang and Loy, Chen Change and Liu, Ziwei},
  journal={International Journal of Computer Vision},
  volume={130},
  number={9},
  pages={2337--2348},
  year={2022},
  publisher={Springer}
}

@inproceedings{zhou2022conditional,
  title={Conditional prompt learning for vision-language models},
  author={Zhou, Kaiyang and Yang, Jingkang and Loy, Chen Change and Liu, Ziwei},
  booktitle={Proceedings of the IEEE/CVF conference on computer vision and pattern recognition},
  pages={16816--16825},
  year={2022}
}

@inproceedings{khattak2023maple,
  title={Maple: Multi-modal prompt learning},
  author={Khattak, Muhammad Uzair and Rasheed, Hanoona and Maaz, Muhammad and Khan, Salman and Khan, Fahad Shahbaz},
  booktitle={Proceedings of the IEEE/CVF Conference on Computer Vision and Pattern Recognition},
  pages={19113--19122},
  year={2023}
}

@inproceedings{khattak2023promptsrc,
  title={Self-regulating prompts: Foundational model adaptation without forgetting},
  author={Khattak, Muhammad Uzair and Wasim, Syed Talal and Naseer, Muzammal and Khan, Salman and Yang, Ming-Hsuan and Khan, Fahad Shahbaz},
  booktitle={Proceedings of the IEEE/CVF International Conference on Computer Vision},
  pages={15190--15200},
  year={2023}
}

@inproceedings{park2024prometar,
  title={Prompt Learning via Meta-Regularization},
  author={Park, Jinyoung and Ko, Juyeon and Kim, Hyunwoo J},
  booktitle={Proceedings of the IEEE/CVF Conference on Computer Vision and Pattern Recognition},
  pages={26940--26950},
  year={2024}
}

@inproceedings{guo2025mmrl,
      title={Mmrl: Multi-modal representation learning for vision-language models},
      author={Guo, Yuncheng and Gu, Xiaodong},
      booktitle={Proceedings of the Computer Vision and Pattern Recognition Conference},
      pages={25015--25025},
      year={2025}
}

@article{gao2024clipadapter,
  title={Clip-adapter: Better vision-language models with feature adapters},
  author={Gao, Peng and Geng, Shijie and Zhang, Renrui and Ma, Teli and Fang, Rongyao and Zhang, Yongfeng and Li, Hongsheng and Qiao, Yu},
  journal={International Journal of Computer Vision},
  volume={132},
  number={2},
  pages={581--595},
  year={2024},
  publisher={Springer}
}

@article{zhang2021tip,
  title={Tip-adapter: Training-free clip-adapter for better vision-language modeling},
  author={Zhang, Renrui and Fang, Rongyao and Zhang, Wei and Gao, Peng and Li, Kunchang and Dai, Jifeng and Qiao, Yu and Li, Hongsheng},
  journal={arXiv preprint arXiv:2111.03930},
  year={2021}
}

@inproceedings{yang2024mma,
  title={Mma: Multi-modal adapter for vision-language models},
  author={Yang, Lingxiao and Zhang, Ru-Yuan and Wang, Yanchen and Xie, Xiaohua},
  booktitle={Proceedings of the IEEE/CVF Conference on Computer Vision and Pattern Recognition},
  pages={23826--23837},
  year={2024}
}

@article{hospedales2021meta,
  title={Meta-learning in neural networks: A survey},
  author={Hospedales, Timothy and Antoniou, Antreas and Micaelli, Paul and Storkey, Amos},
  journal={TPAMI},
  volume={44},
  number={9},
  pages={5149--5169},
  year={2021},
  publisher={IEEE}
}

@inproceedings{ShuXY0ZXM19,
  author       = {Jun Shu and
                  Qi Xie and
                  Lixuan Yi and
                  Qian Zhao and
                  Sanping Zhou and
                  Zongben Xu and
                  Deyu Meng},
  title        = {Meta-Weight-Net: Learning an Explicit Mapping For Sample Weighting},
  booktitle    = {NeurIPS},
  year         = {2019}
}

@inproceedings{BalajiSC18,
  author       = {Yogesh Balaji and
                  Swami Sankaranarayanan and
                  Rama Chellappa},
  title        = {MetaReg: Towards Domain Generalization using Meta-Regularization},
  booktitle    = {NeurIPS},
  year         = {2018}
}

@inproceedings{BechtleMCGRSM20,
  author       = {Sarah Bechtle and
                  Artem Molchanov and
                  Yevgen Chebotar and
                  Edward Grefenstette and
                  Ludovic Righetti and
                  Gaurav S. Sukhatme and
                  Franziska Meier},
  title        = {Meta Learning via Learned Loss},
  booktitle    = {ICPR},
  year         = {2020}
}

@inproceedings{antoniou2018train,
  title={How to train your MAML},
  author={Antoniou, Antreas and Edwards, Harrison and Storkey, Amos},
  booktitle={ICLR},
  year={2019}
}

@inproceedings{munkhdalai2018rapid,
  title={Rapid adaptation with conditionally shifted neurons},
  author={Munkhdalai, Tsendsuren and Yuan, Xingdi and Mehri, Soroush and Trischler, Adam},
  booktitle={ICML},
  year={2018}
}

@inproceedings{lee2019meta,
  title={Meta-learning with differentiable convex optimization},
  author={Lee, Kwonjoon and Maji, Subhransu and Ravichandran, Avinash and Soatto, Stefano},
  booktitle={CVPR},
  year={2019}
}

@inproceedings{hochreiter2001learning,
  title={Learning to learn using gradient descent},
  author={Hochreiter, Sepp and Younger, A Steven and Conwell, Peter R},
  booktitle={ICANN},
  year={2001}
}

@inproceedings{zintgraf2019fast,
  title={Fast Context Adaptation via Meta-Learning},
  author={Zintgraf, Luisa and Shiarli, Kyriacos and Kurin, Vitaly and Hofmann, Katja and Whiteson, Shimon},
  booktitle={ICML},
  pages={7693--7702},
  year={2019}
}

@inproceedings{mishra2018simple,
  title={A simple neural attentive meta-learner},
  author={Mishra, Nikhil and Rohaninejad, Mostafa and Chen, Xi and Abbeel, Pieter},
  booktitle={ICLR},
  year={2018}
}

@inproceedings{sung2018learning,
  title={Learning to compare: Relation network for few-shot learning},
  author={Sung, Flood and Yang, Yongxin and Zhang, Li and Xiang, Tao and Torr, Philip HS and Hospedales, Timothy M},
  booktitle={CVPR},
  pages={1199--1208},
  year={2018}
}

@inproceedings{snell2017prototypical,
  title={Prototypical networks for few-shot learning},
  author={Snell, Jake and Swersky, Kevin and Zemel, Richard},
  booktitle={NeurIPS},
  pages={4077--4087},
  year={2017}
}

@inproceedings{finn2017model,
  title={Model-agnostic meta-learning for fast adaptation of deep networks},
  author={Finn, Chelsea and Abbeel, Pieter and Levine, Sergey},
  booktitle={ICML},
  pages={1126--1135},
  year={2017},
  organization={JMLR. org}
}

@inproceedings{vinyals2016matching,
  title={Matching networks for one shot learning},
  author={Vinyals, Oriol and Blundell, Charles and Lillicrap, Timothy and Wierstra, Daan},
  booktitle={NeurIPS},
  pages={3630--3638},
  year={2016}
}

@inproceedings{santoro2016meta,
  title={Meta-learning with memory-augmented neural networks},
  author={Santoro, Adam and Bartunov, Sergey and Botvinick, Matthew and Wierstra, Daan and Lillicrap, Timothy},
  booktitle={ICML},
  pages={1842--1850},
  year={2016}
}

@inproceedings{munkhdalai2017meta,
    title={Meta Networks},
    author={Tsendsuren Munkhdalai and Hong Yu},
    year={2017},
  booktitle={ICML},
}

@inproceedings{ravi2016optimization,
  title={Optimization as a model for few-shot learning},
  author={Ravi, Sachin and Larochelle, Hugo},
  booktitle={ICLR},
  year={2016}
}

@inproceedings{grant2018recasting,
  title={Recasting gradient-based meta-learning as hierarchical bayes},
  author={Grant, Erin and Finn, Chelsea and Levine, Sergey and Darrell, Trevor and Griffiths, Thomas},
  booktitle={ICLR},
  year={2018}
}

@article{nichol2018first,
  title={On first-order meta-learning algorithms},
  author={Nichol, Alex and Achiam, Joshua and Schulman, John},
  journal={arXiv:1803.02999},
  year={2018}
}

@inproceedings{li2018learning,
  title={Learning to generalize: Meta-learning for domain generalization},
  author={Li, Da and Yang, Yongxin and Song, Yi-Zhe and Hospedales, Timothy},
  booktitle={AAAI},
  year={2018}
}

@article{yao2021meta,
  title={Meta-learning with fewer tasks through task interpolation},
  author={Yao, Huaxiu and Zhang, Linjun and Finn, Chelsea},
  journal={International Conference on Learning Representations},
  year={2021}
}

@inproceedings{koch2015siamese,
  title={Siamese neural networks for one-shot image recognition},
  author={Koch, Gregory},
  booktitle={ICML Workshop},
  year={2015}
}

@inproceedings{hwang2020self,
  title={Self-supervised auxiliary learning with meta-paths for heterogeneous graphs},
  author={Hwang, Dasol and Park, Jinyoung and Kwon, Sunyoung and Kim, KyungMin and Ha, Jung-Woo and Kim, Hyunwoo J},
  booktitle={NeurIPS},
  year={2020}
}

@article{hwang2021self,
  title={Self-supervised auxiliary learning for graph neural networks via meta-learning},
  author={Hwang, Dasol and Park, Jinyoung and Kwon, Sunyoung and Kim, Kyung-Min and Ha, Jung-Woo and Kim, Hyunwoo J},
  journal={IEEE Transactions on Pattern Analysis and Machine Intelligence},
  year={2021}
}

@inproceedings{ko2023meltr,
  title={MELTR: Meta Loss Transformer for Learning to Fine-tune Video Foundation Models},
  author={Ko, Dohwan and Choi, Joonmyung and Choi, Hyeong Kyu and On, Kyoung-Woon and Roh, Byungseok and Kim, Hyunwoo J},
  booktitle={CVPR},
  year={2023}
}

@inproceedings{du2023emo,
  title={EMO: episodic memory optimization for few-shot meta-learning},
  author={Du, Yingjun and Shen, Jiayi and Zhen, Xiantong and Snoek, Cees GM},
  booktitle={Conference on Lifelong Learning Agents},
  pages={1--20},
  year={2023},
  organization={PMLR}
}

@inproceedings{deng2009imagenet,
  title={Imagenet: A large-scale hierarchical image database},
  author={Deng, Jia and Dong, Wei and Socher, Richard and Li, Li-Jia and Li, Kai and Fei-Fei, Li},
  booktitle={2009 IEEE conference on computer vision and pattern recognition},
  pages={248--255},
  year={2009},
  organization={Ieee}
}

@inproceedings{fei2004caltech,
  title={Learning generative visual models from few training examples: An incremental bayesian approach tested on 101 object categories},
  author={Fei-Fei, Li and Fergus, Rob and Perona, Pietro},
  booktitle={2004 conference on computer vision and pattern recognition workshop},
  pages={178--178},
  year={2004},
  organization={IEEE}
}

@inproceedings{parkhi2012pets,
  title={Cats and dogs},
  author={Parkhi, Omkar M and Vedaldi, Andrea and Zisserman, Andrew and Jawahar, CV},
  booktitle={2012 IEEE conference on computer vision and pattern recognition},
  pages={3498--3505},
  year={2012},
  organization={IEEE}
}

@inproceedings{nilsback2008flowers,
  title={Automated flower classification over a large number of classes},
  author={Nilsback, Maria-Elena and Zisserman, Andrew},
  booktitle={2008 Sixth Indian conference on computer vision, graphics \& image processing},
  pages={722--729},
  year={2008},
  organization={IEEE}
}

@inproceedings{krause20133cars,
  title={3d object representations for fine-grained categorization},
  author={Krause, Jonathan and Stark, Michael and Deng, Jia and Fei-Fei, Li},
  booktitle={Proceedings of the IEEE international conference on computer vision workshops},
  pages={554--561},
  year={2013}
}

@article{maji2013aircraft,
  title={Fine-grained visual classification of aircraft},
  author={Maji, Subhransu and Rahtu, Esa and Kannala, Juho and Blaschko, Matthew and Vedaldi, Andrea},
  journal={HAL - INRIA},
  year={2013}
}

@inproceedings{bossard2014food,
  title={Food-101--mining discriminative components with random forests},
  author={Bossard, Lukas and Guillaumin, Matthieu and Van Gool, Luc},
  booktitle={Computer vision--ECCV 2014: 13th European conference, zurich, Switzerland, September 6-12, 2014, proceedings, part VI 13},
  pages={446--461},
  year={2014},
  organization={Springer}
}

@inproceedings{xiao2010sun,
  title={Sun database: Large-scale scene recognition from abbey to zoo},
  author={Xiao, Jianxiong and Hays, James and Ehinger, Krista A and Oliva, Aude and Torralba, Antonio},
  booktitle={2010 IEEE computer society conference on computer vision and pattern recognition},
  pages={3485--3492},
  year={2010},
  organization={IEEE}
}

@inproceedings{cimpoi2014dtd,
  title={Describing textures in the wild},
  author={Cimpoi, Mircea and Maji, Subhransu and Kokkinos, Iasonas and Mohamed, Sammy and Vedaldi, Andrea},
  booktitle={Proceedings of the IEEE conference on computer vision and pattern recognition},
  pages={3606--3613},
  year={2014}
}

@article{helber2019eurosat,
  title={Eurosat: A novel dataset and deep learning benchmark for land use and land cover classification},
  author={Helber, Patrick and Bischke, Benjamin and Dengel, Andreas and Borth, Damian},
  journal={IEEE Journal of Selected Topics in Applied Earth Observations and Remote Sensing},
  volume={12},
  number={7},
  pages={2217--2226},
  year={2019},
  publisher={IEEE}
}

@article{soomro2012ucf101,
  title={UCF101: A dataset of 101 human actions classes from videos in the wild},
  author={Soomro, Khurram  and  Zamir, Amir Roshan  and  Shah, Mubarak},
  journal={CRCV-TR},
  year={2012}
}

@inproceedings{recht2019imagenetv2,
  title={Do imagenet classifiers generalize to imagenet?},
  author={Recht, Benjamin and Roelofs, Rebecca and Schmidt, Ludwig and Shankar, Vaishaal},
  booktitle={International conference on machine learning},
  pages={5389--5400},
  year={2019},
  organization={PMLR}
}

@article{wang2019imagenets,
  title={Learning robust global representations by penalizing local predictive power},
  author={Wang, Haohan and Ge, Songwei and Lipton, Zachary and Xing, Eric P},
  journal={Advances in Neural Information Processing Systems},
  volume={32},
  year={2019}
}

@inproceedings{hendrycks2021imageneta,
  title={Natural adversarial examples},
  author={Hendrycks, Dan and Zhao, Kevin and Basart, Steven and Steinhardt, Jacob and Song, Dawn},
  booktitle={Proceedings of the IEEE/CVF conference on computer vision and pattern recognition},
  pages={15262--15271},
  year={2021}
}

@inproceedings{hendrycks2021imagenetr,
  title={The many faces of robustness: A critical analysis of out-of-distribution generalization},
  author={Hendrycks, Dan and Basart, Steven and Mu, Norman and Kadavath, Saurav and Wang, Frank and Dorundo, Evan and Desai, Rahul and Zhu, Tyler and Parajuli, Samyak and Guo, Mike and others},
  booktitle={Proceedings of the IEEE/CVF international conference on computer vision},
  pages={8340--8349},
  year={2021}
}

@article{holt2004forecasting,
  title={Forecasting seasonals and trends by exponentially weighted moving averages},
  author={Holt, Charles C},
  journal={International journal of forecasting},
  volume={20},
  number={1},
  pages={5--10},
  year={2004},
  publisher={Elsevier}
}

@inproceedings{sun2020test,
  title={Test-time training with self-supervision for generalization under distribution shifts},
  author={Sun, Yu and Wang, Xiaolong and Liu, Zhuang and Miller, John and Efros, Alexei and Hardt, Moritz},
  booktitle={International conference on machine learning},
  pages={9229--9248},
  year={2020},
  organization={PMLR}
}

@inproceedings{wang2021tent,
  title={Tent: Fully Test-Time Adaptation by Entropy Minimization},
  author={Wang, Dequan and Shelhamer, Evan and Liu, Shaoteng and Olshausen, Bruno and Darrell, Trevor},
  booktitle={International Conference on Learning Representations},
  year={2021},
  url={https://openreview.net/forum?id=uXl3bZLkr3c}
}

@inproceedings{liang2020we, 
 title={Do We Really Need to Access the Source Data? Source Hypothesis Transfer for Unsupervised Domain Adaptation}, 
 author={Liang, Jian and Hu, Dapeng and Feng, Jiashi}, 
 booktitle={International Conference on Machine Learning (ICML)},  
 pages={6028--6039},
 year={2020}
}

@article{liang2021source,  
 title={Source Data-absent Unsupervised Domain Adaptation through Hypothesis Transfer and Labeling Transfer}, 
 author={Liang, Jian and Hu, Dapeng and Wang, Yunbo and He, Ran and Feng, Jiashi},   
 journal={IEEE Transactions on Pattern Analysis and Machine Intelligence (TPAMI)},
 year={2021}, 
 note={In Press}  
}

@inproceedings{sun2024learning,
  title={Learning to (learn at test time): Rnns with expressive hidden states},
  author={Sun, Yu and Li, Xinhao and Dalal, Karan and Xu, Jiarui and Vikram, Arjun and Zhang, Genghan and Dubois, Yann and Chen, Xinlei and Wang, Xiaolong and Koyejo, Sanmi and others},
  booktitle = {ICML},
  year={2025}
}

\clearpage
\appendix
\section{Related Work}
\label{sec:related_work}

\paragraph{Vision-language models (VLMs).}
Contrastive Language–Image Pretraining (CLIP)~\cite{radford2021learning} is a vision-language model demonstrating strong zero-shot classification via alignment of test images with handcrafted prompts.
However, adapting CLIP to downstream tasks through full fine-tuning often degrades its zero-shot generalization.
To mitigate this, various parameter-efficient fine-tuning (PEFT) methods have been proposed.
CoOp~\cite{zhou2022learning} formulates prompt tuning by introducing learnable textual prompts,
while CoCoOp~\cite{zhou2022conditional} introduces sample-specific prompts to reduce overfitting to base classes.
MaPLe~\cite{khattak2023maple} adopts a prefix-tuning approach by injecting learnable prefixes into shallow layers of both encoders, preserving handcrafted prompts.
Building upon this, PromptSRC~\cite{khattak2023promptsrc} incorporates self-regulating losses to constrain prefix learning.
ProMetaR~\cite{park2024prometar} leverages meta-learning to optimize these regularization terms, enhancing robustness across diverse tasks.
MMRL~\cite{guo2025mmrl} diverges by inserting learnable tokens into deep encoder layers to better retain general representations.
In parallel, adapter-based methods provide an alternative PEFT paradigm. 
CLIP-Adapter~\cite{gao2024clipadapter} appends lightweight MLP adapters,
while Tip-Adapter~\cite{zhang2021tip} leverages a cache-based design.
MMA~\cite{yang2024mma} introduces shared-space adapters in deep layers to enhance multimodal integration.
While these methods primarily target few-shot settings, our work focuses on test-time tuning to improve zero-shot performance without access to training data.

\paragraph{Test-time adaptation (TTA).}
TTA addresses distribution shifts by adapting a source-pretrained model to the target domain at test time.
Recent studies have extended TTA to VLMs along two lines:
Training-required approaches fine-tune a subset of model parameters at test time.
TPT~\cite{shu2022test} proposes test-time prompt tuning by adapting CoOp~\cite{zhou2022learning}'s learnable prompts  via augmented views of test samples.
DiffTPT~\cite{feng2023diverse} improves view diversity using diffusion~\cite{rombach2022diffusion}-based generation.
PromptAlign~\cite{abdul2023align} adopts test-time prefix tuning to align feature distributions, aligning feature distributions to adapt MaPLe~\cite{khattak2023maple}.
Training-free approaches refine predictions via non-parametric mechanisms.
TDA~\cite{karmanov2024efficient} constructs positive and negative caches from previous test samples for test-time inference.
BoostAdapter~\cite{zhang2024boostadapter} improves this by incorporating sample-specific augmented views into the cache.
Our method follows the training-required paradigm, addressing limitations of prompt tuning.

\paragraph{Meta-learning.}
Meta-learning, or ``learning to learn'', focuses on rapidly adapting models to new tasks by leveraging prior knowledge, as described by Hospedales \textit{et al.}~\cite{hospedales2021meta}. 
This approach has been applied across various domains, including the design of loss functions~\cite{ShuXY0ZXM19,BalajiSC18,BechtleMCGRSM20}, the development of task-specific initializations~\cite{finn2017model}, and the enablement of few-shot learning~\cite{koch2015siamese,sung2018learning,snell2017prototypical}. 
Meta-learning techniques are typically classified into metric-based~\cite{snell2017prototypical,sung2018learning,vinyals2016matching,lee2019meta}, memory-based~\cite{mishra2018simple,santoro2016meta,munkhdalai2017meta,munkhdalai2018rapid,hochreiter2001learning, du2023emo}, and gradient-based methods~\cite{ravi2016optimization,grant2018recasting,nichol2018first,li2018learning}. Since the emergence of Model-Agnostic Meta-Learning (MAML)~\cite{finn2017model}, gradient-based meta-learning has received substantial attention; however, these approaches often encounter meta-overfitting issues, especially with limited meta-training tasks~\cite{antoniou2018train,yao2021meta,zintgraf2019fast,hwang2020self,hwang2021self,ko2023meltr}. 
In light of these challenges, we introduce a meta-learning test-time adaptation framework that dynamically adapts VLMs to diverse testing scenarios.

\section{Additional Implementation Details}
\paragraph{Datasets.}
Following TPT~\cite{shu2022test}, we conduct experiments on two zero-shot generalization benchmarks.
For \textit{domain generalization}, we evaluate across four out-of-distribution (OOD) variants of ImageNet~\cite{deng2009imagenet}:
ImageNet-V2~\cite{recht2019imagenetv2},
ImageNet-Sketch~\cite{wang2019imagenets},
ImageNet-A~\cite{hendrycks2021imageneta}
and ImageNet-R~\cite{hendrycks2021imagenetr}.
For \textit{cross-dataset evaluation}, we conduct evaluations on ten image classification datasets:
Caltech101~\cite{fei2004caltech},
OxfordPets~\cite{parkhi2012pets},
OxfordFlowers~\cite{nilsback2008flowers},
StanfordCars~\cite{krause20133cars},
FGVC-Aircraft~\cite{maji2013aircraft},
Food101~\cite{bossard2014food},
SUN397~\cite{xiao2010sun},
DTD~\cite{cimpoi2014dtd},
EuroSAT~\cite{helber2019eurosat}
and UCF101~\cite{soomro2012ucf101}.

\paragraph{Baselines.}
We evaluate MetaTPT against several baselines:
(1) Zero-shot CLIP~\cite{radford2021learning}.
(2) Few-shot prompt learning methods: CoOp~\cite{zhou2022learning}, MaPLe~\cite{khattak2023maple} and MMRL~\cite{guo2025mmrl}.
(3) Test-time prompt tuning methods: TPT~\cite{shu2022test} and PromptAlign~\cite{abdul2023align}.

\section{Additional Ablation Studies}
\label{sec:appendix}
\begin{figure}[t]
    \centering
    \includegraphics[width=\linewidth]{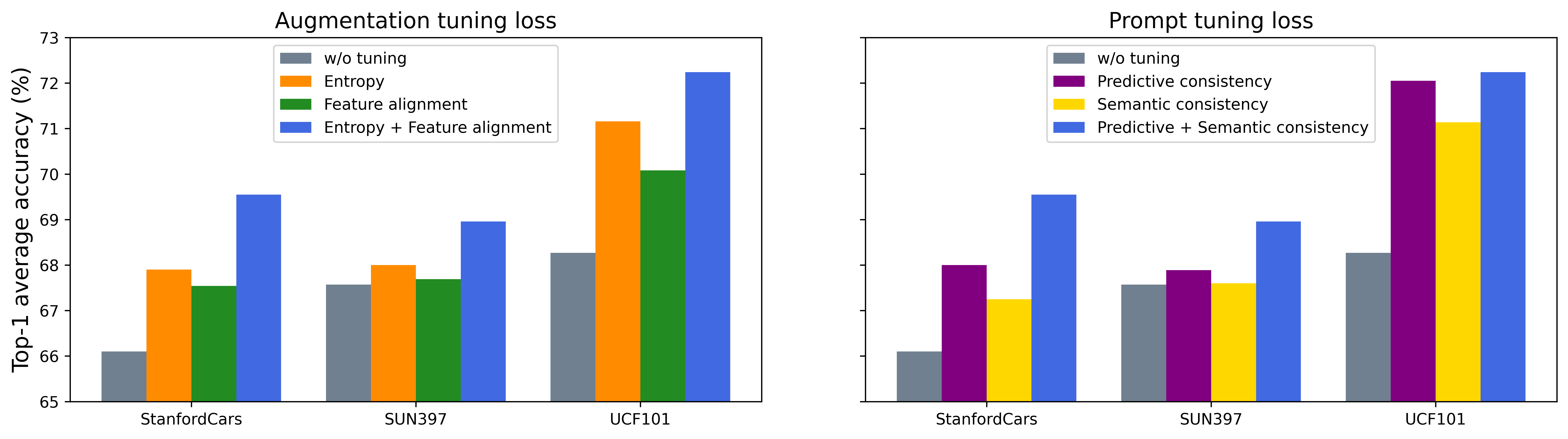}
    \caption{\textbf{Ablation of Loss Components} on augmentation tuning loss $\mathcal{L}_\textrm{inner}$ and prompt tuning loss $\mathcal{L}_\textrm{outer}$, evaluated on StanfordCars, SUN397 and UCF101.}
    \label{fig:loss}
\end{figure}

\paragraph{Loss ablation.}
Figure~\ref{fig:loss} depicts a loss ablation study conducted on StanfordCars, SUN397, and UCF101, comparing different loss configurations for augmentation tuning (left) and prompt tuning (right).
For augmentation tuning, we evaluate four settings: MMRL~\cite{guo2025mmrl} without tuning, tuning $\mathbf{\Phi}$ with only entropy loss $H$, only feature alignment loss $\mathcal{L}_{\text{dis}}$, and our proposed combination of both.
While individually applying entropy or feature alignment loss improves performance relative to MMRL, their joint optimization consistently achieves superior results.
Similarly, for prompt tuning, we assess MMRL without tuning, tuning $\mathbf{\Theta}$ with only predictive consistency $\mathcal{L}_{ce}$, only semantic consistency $\mathcal{L}_{\text{dis}}$, and our combined predictive and semantic consistency losses. The trend is consistent: combining predictive and semantic consistency losses achieves superior adaptation compared to using either loss individually.

\begin{figure}[t]
    \centering
    \includegraphics[width=\linewidth]{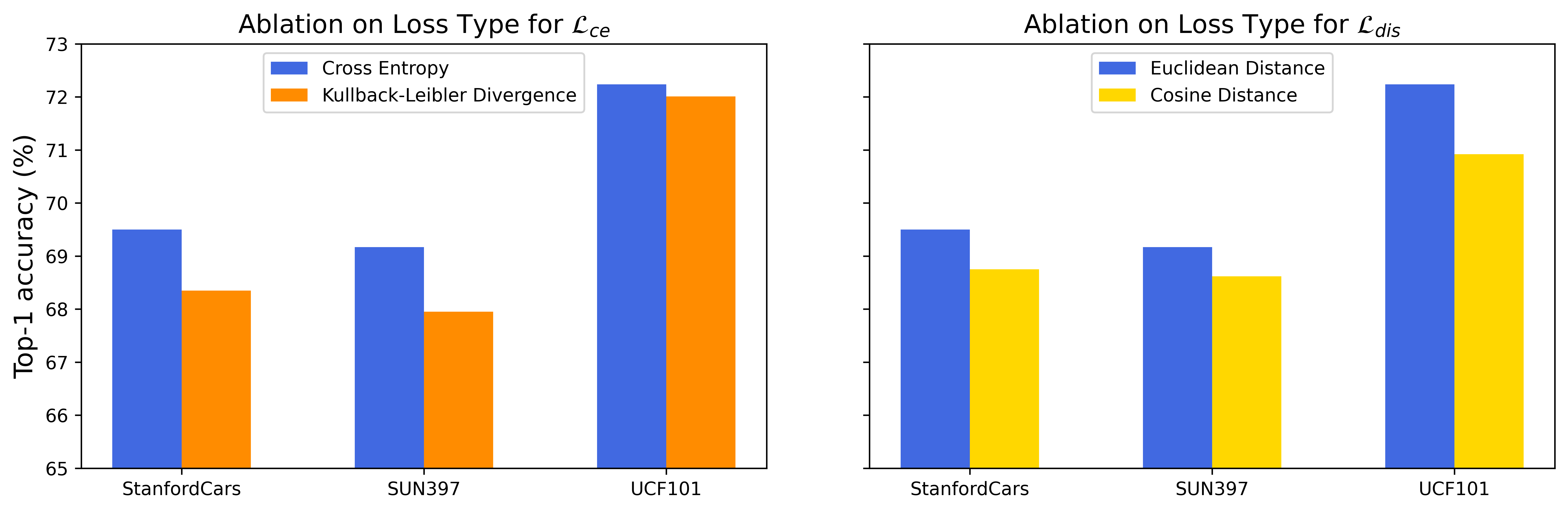}
    \caption{\textbf{Ablation study on loss types for predictive consistency loss $\mathcal{L}_{ce}$ and feature alignment loss $\mathcal{L}_\textrm{dis}$}, evaluated on StanfordCars, SUN397 and UCF101.}
    \label{fig:loss_type}
\end{figure}

\paragraph{Ablation on loss types in test-time prompt tuning.}
Figure~\ref{fig:loss_type} evaluates the impact of different loss functions on test-time prompt tuning (Section~\ref{sec: outer}).
For predictive consistency $\mathcal{L}_{ce}$, we compare cross-entropy and Kullback–Leibler (KL) divergence. Cross-entropy leads to more reliable alignment of soft predictions than KL divergence, improving output stability. 
For semantic consistency $\mathcal{L}_\textrm{dis}$, we evaluate Euclidean distance and cosine distance. Euclidean distance outperforms cosine distance by achieving stronger semantic alignment between features.
Taken together, these findings suggest that combining cross-entropy with Euclidean distance provides the most effective objective for test-time prompt tuning.

\begin{figure}[t]
    \centering
    \begin{minipage}[h]{0.48\linewidth}
        \centering
        \includegraphics[width=\linewidth]{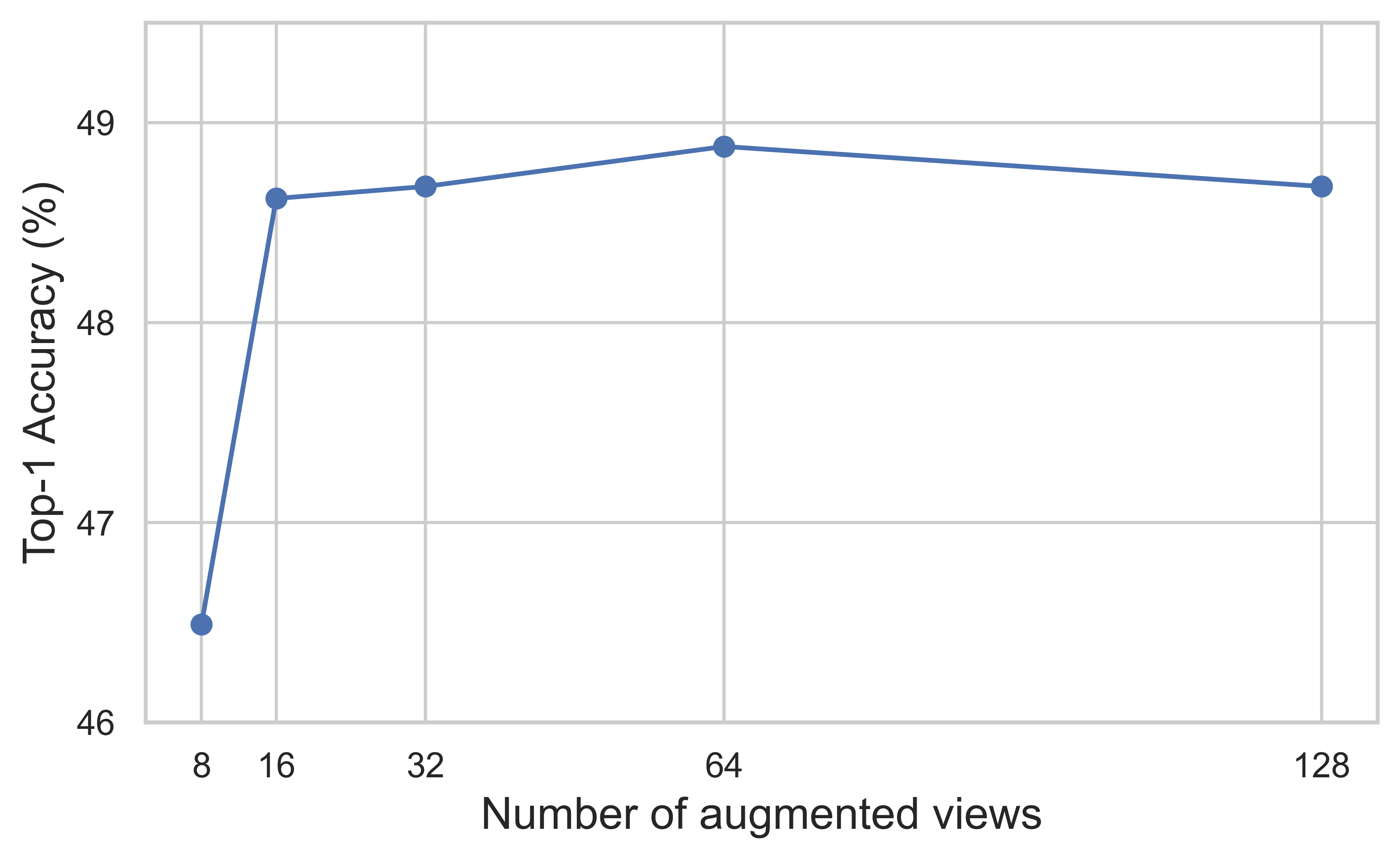}
        \caption{\textbf{Ablation study on the number of augmented views $N$}, evaluated on DTD.}
        \label{fig:batch_size}
    \end{minipage}
    \hfill
    \begin{minipage}[h]{0.48\linewidth}
        \centering
        \includegraphics[width=\linewidth]{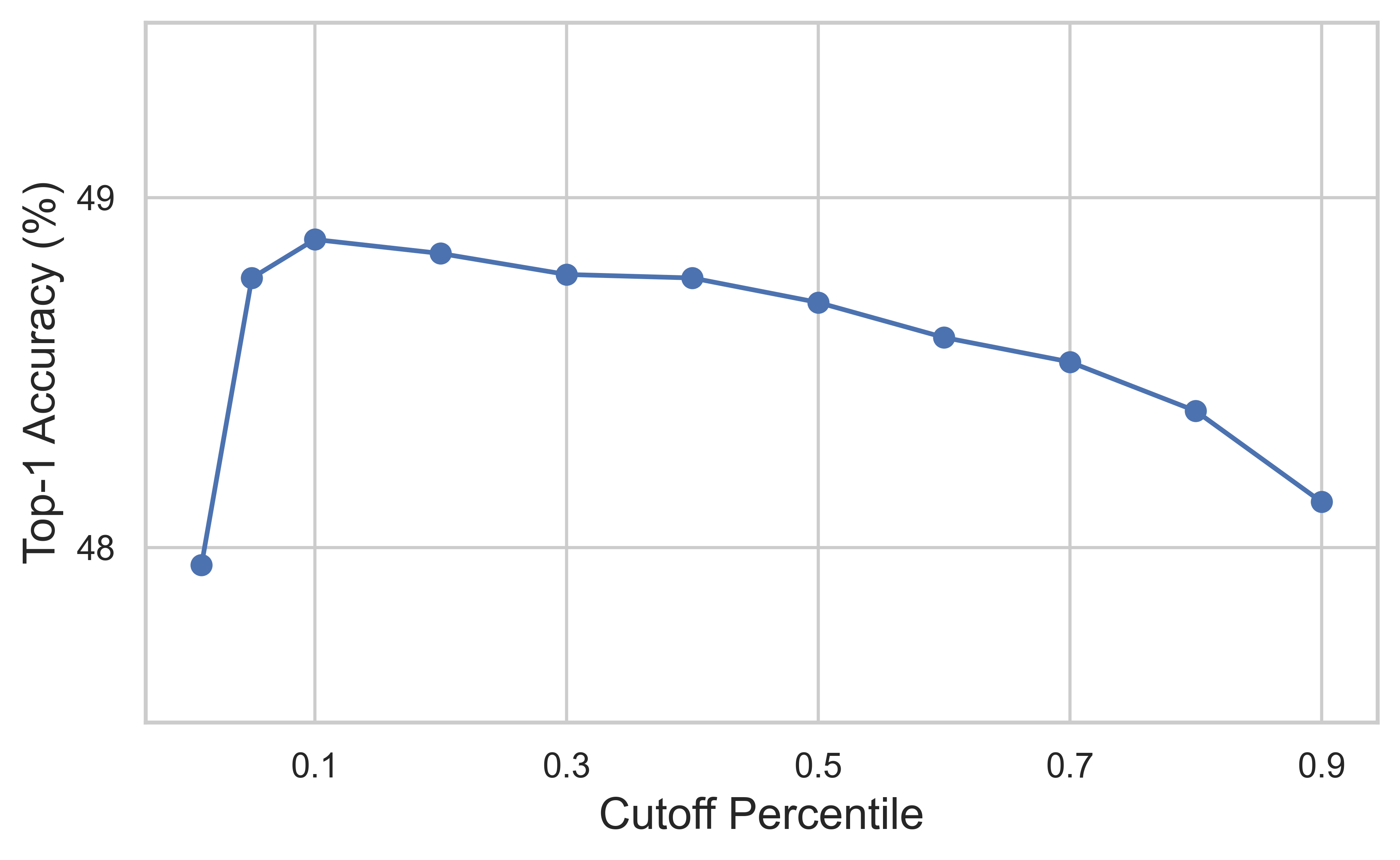}
        \caption{\textbf{Ablation study on the cutoff percentile $\rho$ in confidence selection}, evaluated on DTD.}
        \label{fig:rho}
    \end{minipage}
\end{figure}

\paragraph{Ablation on numbers of augmented views.}
Figure~\ref{fig:batch_size} reports an ablation study on the number of augmented views on model performance. Performance improves steadily as the number of views increases from 8 to 64, indicating that greater augmentation diversity benefits representation learning. Beyond 64 views, gains become marginal while computational overhead rises notably. These results highlight 64 views as an effective trade-off between accuracy and computational efficiency.

\paragraph{Ablation on confidence selection threshold.}
Figure~\ref{fig:rho} reports an ablation study on the confidence selection threshold $\rho$ defined in Eq.~(\ref{eq: selection}), evaluated across a range of values from 0.01 to 0.9. Performance peaks at $\rho = 0.1$, indicating an optimal balance between retaining sufficient samples and maintaining label quality. Thresholds below 0.1 introduce noisy pseudo-labels, degrading performance, while higher thresholds excessively reduce the training signal by filtering out too many samples.

\paragraph{Results of different CLIP backbones.}
Table~\ref{tab:b32_cd} and Table~\ref{tab:b32_dg} report the performance of MetaTPT using the ViT-B/32 backbone on cross-dataset and domain generalization benchmarks, respectively. On the cross-dataset benchmark (Table~\ref{tab:b32_cd}), MetaTPT achieves the highest average accuracy of 65.57\%, surpassing strong baselines including MaPLe~\cite{khattak2023maple}, PromptAlign~\cite{abdul2023align}, and MMRL~\cite{guo2025mmrl}. Improvements are consistent across most datasets, demonstrating the effectiveness of MetaTPT in handling distribution shifts. On the domain generalization benchmark (Table~\ref{tab:b32_dg}), which includes four ImageNet variants, MetaTPT again outperforms all competitors, reaching an average accuracy of 54.55\%. These results demonstrate that MetaTPT consistently enhances generalization performance across domain-shift settings, highlighting its effectiveness regardless of backbone capacity.

\begin{table}[h]
\centering
\caption{\textbf{Comparison of MetaTPT in Cross-Dataset Evaluation on ViT-B/32}, conducted across ten datasets.}
\label{tab:b32_cd}
\resizebox{\textwidth}{!}{%
\begin{tabular}{l|cccccccccc|c}
\toprule
& \rotatebox{90}{Caltech} & \rotatebox{90}{Pets} & \rotatebox{90}{Cars} & \rotatebox{90}{Flowers} & \rotatebox{90}{Food101} & \rotatebox{90}{Aircraft} & \rotatebox{90}{SUN397} & \rotatebox{90}{DTD} & \rotatebox{90}{EuroSAT} & \rotatebox{90}{UCF101} & \rotatebox{90}{\textit{Average}}  \\ \midrule
MaPLe~\cite{khattak2023maple} & 92.50 & 88.13 & 59.93 & 65.33 & 81.00 & 17.53 & 65.00 & 41.70 & 40.80 & 63.63 & 61.56 \\
MaPLe + TPT~\cite{shu2022test} & 91.44 & 88.47 & 59.35 & 66.08 & 82.08 & 18.71 & 66.07 & 40.01 & 39.67 & 61.63 & 61.35 \\
PromptAlign~\cite{abdul2023align} & 92.10 & 88.44 & 63.48 & 66.14 & 82.07 & 18.76 & 66.08 & 42.54 & 39.68 & 65.57 & 62.49 \\
\midrule
MMRL~\cite{guo2025mmrl} & 92.40 & 88.70 & 60.78 & 66.94 & 80.59 & 20.33 & 65.44 & 46.93 & \textbf{48.86} & 65.63 & 63.66 \\
\rowcolor[HTML]{E1E1E1} 
\textbf{MMRL + MetaTPT} & \textbf{93.12} & \textbf{89.79} & \textbf{64.41} & \textbf{68.47} & \textbf{82.56} & \textbf{22.75} & \textbf{68.05} & \textbf{50.24} & 47.10 & \textbf{69.26} & \textbf{65.57} \\
\bottomrule
\end{tabular}%
}
\end{table}

\begin{table}[h]
\centering
\tabcolsep=0.4cm
\caption{\textbf{Comparison of MetaTPT in Domain Generation on ViT-B/32}, conducted on four ImageNet variants.}
\label{tab:b32_dg}
\resizebox{\textwidth}{!}{%
\begin{tabular}{l|cccc|c}
\toprule
 & ImageNet-V2 & ImageNet-Sketch & ImageNet-A & ImageNet-R & \textit{Average} \\
 \midrule
MaPLe~\cite{khattak2023maple} & 57.63 & 42.15 & 32.12 & 67.64 & 49.89 \\
MaPLe + TPT~\cite{shu2022test} & 60.01 & 43.77 & 37.52 & 71.11 & 53.10 \\
PromptAlign~\cite{abdul2023align} & 60.43 & 44.24 & 38.02 & 71.44 & 53.53 \\
\midrule
MMRL~\cite{guo2025mmrl} & 58.22 & 42.59 & 32.09 & 68.26 & 50.29 \\
\rowcolor[HTML]{E1E1E1} 
\textbf{MMRL + MetaTPT} & \textbf{61.31} & \textbf{45.63} & \textbf{38.65} & \textbf{72.54} & \textbf{54.53} \\
\bottomrule
\end{tabular}%
}
\end{table}

\section{Additional Discussions}
\label{sec:limit}
One limitation of MetaTPT is its sample-specific optimization overhead at test time. While we adopt amortized and parallelizable augmentation updates, the dual-loop structure and per-sample affine parameters still introduce extra latency compared to training-free or batch-level adaptation methods. Future work could explore more efficient augmentation parameterizations, shared initialization across similar samples, or learning lightweight controllers to modulate augmentation without per-sample gradient steps.

\end{document}